\documentclass[5]{elsarticle}

\usepackage{multirow}
\usepackage{hyperref}
\usepackage{graphicx}
\usepackage{tabularx,booktabs}
\usepackage{dingbat}
\usepackage{diagbox}
\newcolumntype{C}{>{\centering\arraybackslash}X}
\usepackage{xcolor}
\usepackage{times}
\usepackage{epsfig}
\usepackage{graphicx}
\usepackage{amsmath}
\usepackage{amssymb}
\usepackage{csquotes}
\usepackage{algorithm}
\usepackage{algorithmicx}
\usepackage{algpseudocode}
\usepackage{mathtools}
\usepackage{amsfonts}
\usepackage{csquotes}
 
\DeclareMathOperator*{\argmax}{argmax}

\algdef{SE}[DOWHILE]{Do}{doWhile}{\algorithmicdo}[1]{\algorithmicwhile\ #1}
\usepackage{tikz}
\usepackage{url}

\usepackage{pifont}
\newcommand{\xmark}{\ding{55}}%

\biboptions{compress}

\begin{document}

\begin{frontmatter}

\title{Deep Convolutional Correlation Iterative Particle Filter for Visual Tracking}

\author[1]{Reza {Jalil Mozhdehi}\corref{cor1}} 
\ead{reza.jalilmozhdehi@marquette.edu}

\author[2]{Henry {Medeiros}\fnref{fn2}}
\ead{hmedeiros@ufl.edu}

\address[1]{Department of Electrical and Computer Engineering, Marquette University, WI, USA}

\address[2]{Department of Agricultural and Biological Engineering, University of Florida, FL, USA}

\cortext[cor1]{Corresponding author;}

\fntext[fn2]{Part of this work was performed while the author was with the Department of Electrical and Computer Engineering, Marquette University.}

\begin{abstract}
This work proposes a novel framework for visual tracking based on the integration of an iterative particle filter, a deep convolutional neural network, and a correlation filter. The iterative particle filter enables the particles to correct themselves and converge to the correct target position. We employ a novel strategy to assess the likelihood of the particles after the iterations by applying K-means clustering. Our approach ensures a consistent support for the posterior distribution. Thus, we do not need to perform resampling at every video frame, improving the utilization of prior distribution information. Experimental results on two different benchmark datasets show that our tracker performs favorably against state-of-the-art methods.
\end{abstract}


\end{frontmatter}

\section{Introduction}

Visual tracking is a challenging computer vision problem in which the size and location of a specific target are provided in the first video frame, and the target is then followed in subsequent frames by estimating its size and position. What makes the problem particularly challenging is the fact that the appearance of the target may change significantly in scenarios such as those involving partial occlusion or deformation. The introduction of deep convolutional neural networks (CNNs) \citep{NIPS2012,SimonyanICLR2015} to extract target features for visual tracking was a turning point in the design of tracking algorithms \citep{DLT,CNNSVMICML2015}. These convolutional features, in conjunction with correlation filters such as those proposed in \citep{dai2019visual,zhang2018ECCV,qi2016hedged,HCFT,mozhdehideep,FU2020,RAJU2021}, significantly improve tracking performance in comparison with traditional correlation filters based on hand-crafted features \citep{ZhangECCV2014, ZhongTIP2014, Struck, TLD}. The methods described in \citep{mozhdehideep} and \citep{MCPF2017CVPR} were the first approaches to demonstrate the effectiveness of employing particle filters in conjunction with correlation-convolution trackers. In these methods, particle filters are used to sample several image patches, which are then processed by a CNN. The weight of each sample is calculated by applying a correlation filter to the convolutional features

\begin{figure}
\centering
    \includegraphics[width=.4\textwidth]{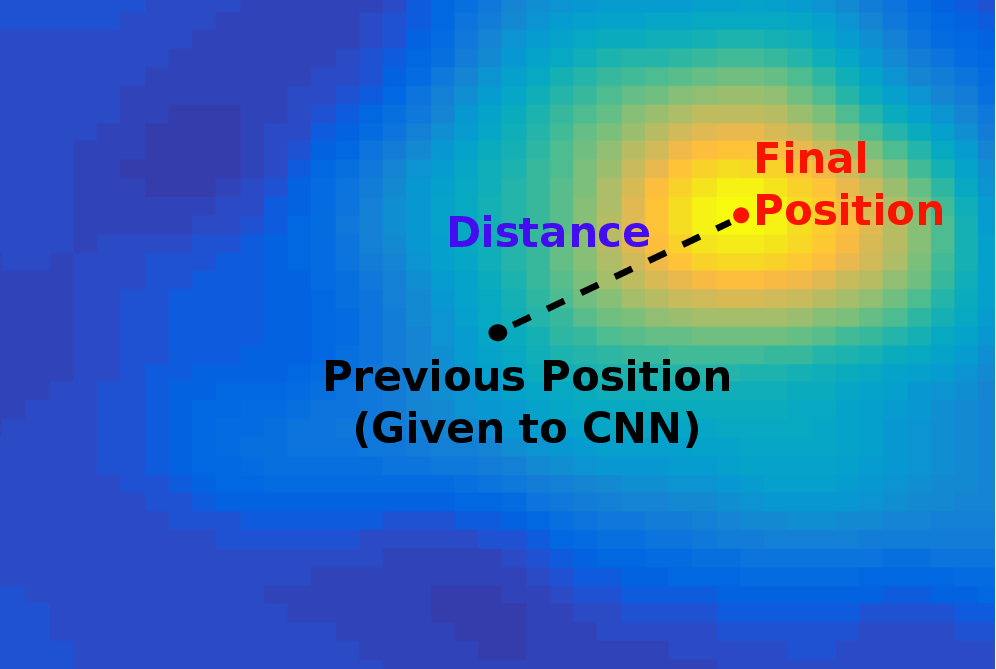}
\caption{Example of a correlation map of a given frame with respect to the previous target position.}
\label{fig:1new}
\end{figure}

In this paper, we propose a deep convolutional correlation iterative particle filter (D2CIP) tracker. D2CIP is an extension of our previous visual trackers \citep{mozhdehideep,mozhdehideep2,mozhdehideep3,mozhdehideep4}, which represent a new class of tracking algorithms that integrate Sequential Monte Carlo strategies with correlation-convolution techniques. Our proposed tracker uses multiple particles as the inputs to a CNN \citep{SimonyanICLR2015} and then applies the correlation filter used in the ECO tracker \citep{ECO} to generate the correlation map of each particle. As Fig. \ref{fig:1new} illustrates, trackers based on correlation filters attempt to determine the new position of the target by analyzing the displacement between the center of the correlation map, which corresponds to the previous target position, and the new peak in the map. More specifically, the correlation between the target model generated at  previous frames and convolutional features extracted from the current frame are used to determine the new target position. Larger displacements between the previous target position and the peak of the correlation map lead to a degradation in the quality of the correlation map, since the corresponding convolutional features are less similar to the target model. Our proposed method differs from previous approaches in that it decreases this distance for each particle through an iterative procedure. At each iteration, the particles approach the target location and an improved correlation map is computed. To our knowledge, iterative particle filters have not been used in conjunction with CNNs and correlation filters before.  

\begin{figure*}[ht]
     \centering
\includegraphics[width=.79\textwidth]{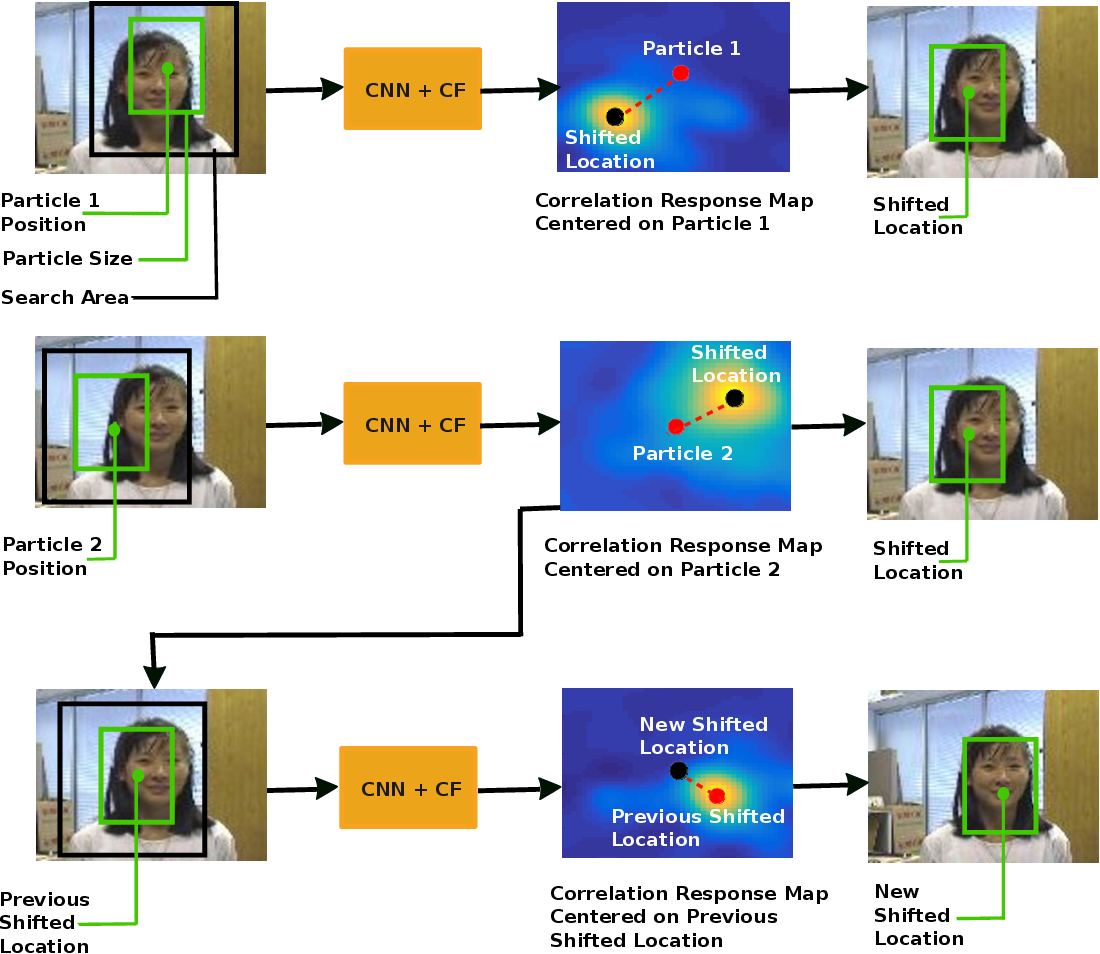}
\centering
\begin{small}
\caption {The top and middle rows illustrate how two distinct particles with different weights converge to the same shifted location. Two different patches centered in particles $1$ and $2$ are given to the CNN and correlation filter to generate their correlation response maps. Each of these particles is then shifted to the peak of its corresponding correlation response map. As shown, their shifted locations are identical. Thus, this location is associated with two different weights in the posterior distribution because the correlation maps corresponding to the two particles are different. The middle and bottom rows show how the particle and its shifted location may generate different correlation maps. In the bottom row, a patch centered in the shifted location of particle $2$ is generated. This patch results in a different correlation map at a different shifted location in comparison with the patch corresponding to particle $2$. Thus, if the shifted location of particle $2$ is used in the computation of the posterior distribution, its weight should be calculated based on the correlation map on the bottom row instead of the one on the middle row.} 
\label{fig:2problem}
\end{small}
\end{figure*}

\begin{figure*}[t]
     \centering    
\includegraphics[trim=0 2 0 0, clip, width=0.9\textwidth]{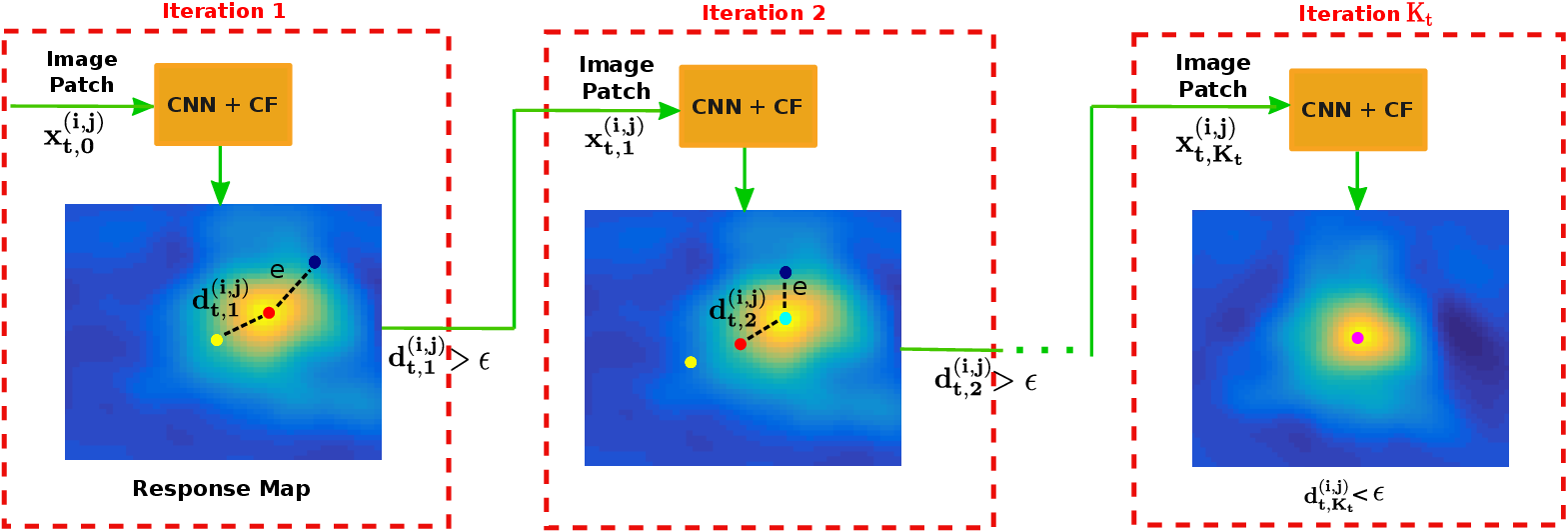}
\caption {Illustration of the proposed iterative particle position refinement. In the first iteration, particle $x_{t,0}^{(i,j)}$ shown by the yellow point is given to the CNN and the correlation filter to calculate its response map. Because the displacement $d_{t,1}^{(i,j)}$ between the estimated position and the particle is higher than $\epsilon$, the particle needs to be refined. In this scenario, the estimated position is not accurate and there is an error $e$ between the peak of the correlation map and the ground truth position (black point). The red point is then considered the new particle $x_{t,1}^{(i,j)}$ for the second iteration. The cyan point shows the estimated position in the second iteration, which needs further refinement despite the reduction in the error $e$. The purple point represents the position at the $K_{t}$-th iteration, which does not need to be refined because $d_{t,K_{t}}^{(i,j)}<\epsilon$.} 
\label{fig:4iterative}
\end{figure*}

The second major contribution of our work is a novel target state estimation strategy. In our previous particle-correlation trackers  \citep{mozhdehideep,MCPF2017CVPR,mozhdehideep2,mozhdehideep3,mozhdehideep4} and similar state-of-the-art methods \citep{Zhang2018,Redetection}, assessing the likelihood of the particles is challenging because many particles may be in close proximity to one another. In the framework proposed in this paper, the particles converge to a few locations after a series of iterations. Thus, we propose a novel method based on particle clustering and convergence consistency to evaluate the final particle locations. This novel method enables our proposed tracker to overcome challenges associated with multi-modal likelihood distributions. Additionally, existing particle-correlation trackers must perform resampling at every frame because shifting the particles to the peak of the correlation maps changes the support of the posterior distribution. Our iterative particle filter overcomes this issue and hence does not disregard information from prior samples. We tested our tracker on the LaSOT dataset \citep{LaSOT}, the TREK-150 dataset \citep{TREK150}, and the Visual Tracker Benchmark v1.1 beta (OTB100) \citep{OTB}. The results show that our tracker outperforms several state-of-the-art methods.

\section{Related work}
The successful application of deep convolutional neural networks to object detection tasks \cite{Salient2020, FPN2017, Girshick2016, Ren2015, Siddique, ISVC2018Bashiri} has led to an increased interest in the utilization of such networks for visual tracking applications. Most CNN-based tracking algorithms use a CNN to examine image patches and determine the likelihood that a particular patch corresponds to the target. Li et al. \cite{Li2016DeepTrack} presented a tracker which samples image patches from the region surrounding the previous target position and uses multiple image cues as inputs to a CNN.  The authors later employed Bagging \cite{Bagging2014} to improve the robustness of their online network weight update process \cite{Bagging2016}. The multi-domain network (MDNet) tracker \cite{MDNet2016CVPR} samples image patches at multiple positions and scales to account for target size variations. MDNet uses three convolutional layers to extract common target features and several domain- (or target-) specific fully connected layers to differentiate between a certain target category and the background. SANet \cite{DBLP:journals/corr/FanL16a} extends the MDNet architecture by employing a recurrent neural network (RNN) to predict the target position. However, much of the performance of MDNet and SANet is due to the fact that they are trained by utilizing the benchmark datasets on which they are evaluated.

One effective mechanism to determine the similarity between an image patch and the target are correlation filters \citep{Choi_2016_CVPR,tang2015multi,SRDCF}. Trackers based on correlation filters measure the correlation between the target model and an image patch in the frequency domain and are agnostic to the features used to represent the targets. By employing correlation filters on the convolutional features generated by multiple layers of a deep CNN, HCFT \citep{HCFT} shows substantial performance improvement in comparison with other visual trackers. Later, Qi et al. \citep{qi2016hedged} introduced HDT, which augments HCFT with a hedging algorithm to weigh the outputs of the convolutional layers, better leveraging semantic information and improving tracking results \citep{1212}. Instead of considering convolutional layers independently, Danelljan et al. proposed C-COT \citep{C-cot}, which employs a continuous fusion method among multiple convolutional layers and uses a joint learning framework to leverage different spatial resolutions. The authors later addressed C-COT's problems of computational complexity and model over-fitting in the ECO algorithm \citep{ECO}. Their factorized convolution operator and their novel model update method decrease the number of parameters in the model and  improve tracking speed and robustness.

Recently, several methods have been proposed to improve the performance of corre\-lation-based trackers. One strategy entails combining different types of features and constructing multiple correlation-based experts \citep{WangMultiExperts}.   Spatial-temporal information can also be used to address unwanted boundary effects in correlation trackers by spatially penalizing the filter coefficients \citep{SRDCF}. To address the additional computational costs associated with such strategies, in contrast to methods that train the model using samples from multiple frames, \cite{STRCF2018} update the correlation model using samples from the current frame and the previously learned correlation filter. \cite{flowtrack} further extend such strategies through a spatial-temporal attention mechanism that uses optical flow information in consecutive frames. Finally, \cite{DRT} use an approach based on reliability information \citep{Tang2020Reliability}, which performs real-time tracking by estimating the importance of sub-regions within the correlation filter.

Particle filters provide an effective and general framework for improving the performance of correlation-based trackers. However, using particle filters in conjunction with correlation filters also involves some challenges. As shown in \citep{mozhdehideep,Zhang2018}, particle-correlation trackers use the sum of the elements of the correlation maps as the weights of the particles. However, in challenging situations, such as in the presence of occlusions or target deformations, the correlation maps are not reliable and generate weights that do not reflect the similarity between the target model and the image patch under consideration. Additionally, particle-correlation trackers generally estimate the target state based on the particle with the maximum weight \citep{mozhdehideep,Zhang2018,MCPF2017CVPR}, which is not always an accurate method because many particles may have similar weights. Furthermore, the aforementioned correlation-particle trackers perform resampling at every frame  and consequently disregard previous particle information. As shown in \citep{Zhenhua2015Iterative}, iterative particle filters can improve sampling and lead to more distinctive particle likelihood models. However, such methods have been not used in conjunction with correlation-convolution trackers so far. 

\section{The change of support problem in convolution-correlation particle filters}\label{sec:3}

Recursive Bayesian estimation algorithms attempt to determine the distribution of the target state $x_t$ given a set of observations $y_{0:t}= \left\{y_0, y_1, \ldots, y_t \right\}$ using Bayes rule
\begin{equation} \label{eq:bayes}
p(x_t|y_{0:t})=\frac{p(y_t|x_t,y_{0:t-1})p(x_t|y_{0:t-1})}{p(y_t|y_{0:t-1})}.
\end{equation}
Since analytical solutions to Eq. \ref{eq:bayes} are only available for very specific classes of problems (e.g., when all the distributions are normal), particle filters employ a Monte Carlo strategy to approximate the distribution of the target state. That is, a set of particles $\left\{x_t^{(i)}\right\}_{i=0}^{N}$ is sampled from a proposal distribution $q(x^{(i)}_{t}|x^{(i)}_{t-1},y_{t})$ and the contribution of each sample is weighed according to 
a likelihood function $p(y_{t}|x^{(i)}_{t})$ and a  transition distribution $p(x^{(i)}_{t}|x^{(i)}_{t-1})$. In a particle filter, the weight of each particle is calculated by
\begin{equation} \label{eq:1}
     \omega^{(i)}_{t} \propto \omega^{(i)}_{t-1} \dfrac {p(y_{t}|x^{(i)}_{t}) p(x^{(i)}_{t}|x^{(i)}_{t-1})}{q(x^{(i)}_{t}|x^{(i)}_{t-1},y_{t})},
\end{equation}
and posterior target state distribution is approximated by 
\begin{equation} \label{eq:2}
p(x_{t}|y_{t}) \approx \sum_{i=1}^{N} \bar{\omega}^{(i)}_{t} \delta(x_{t}-x^{(i)}_{t}),
\end{equation}
where $\bar{\omega}^{(i)}_{t}$ are the normalized particle weights. For additional details, we refer the reader to \citep{Tutorial}.

Particle filters used in correlation trackers such as  \citep{mozhdehideep,MCPF2017CVPR,mozhdehideep2,mozhdehideep3,mozhdehideep4,Zhang2018,Redetection} sample particles from the transition distribution. Thus, $q(x^{(i)}_{t}|x^{(i)}_{t-1},y_{t})=p(x^{(i)}_{t}|x^{(i)}_{t-1})$. Additionally, they resample particles at every frame, which  removes the dependency on previous weights. Hence, the weight of each particle in these trackers is 
\begin{equation} \label{eq:3}
     \omega^{(i)}_{t} \propto p(y_{t}|x^{(i)}_{t}).
\end{equation}
In these trackers, the particles are shifted to the peak of the correlation maps and the posterior distribution is then calculated by the particles' weights and the corresponding shifted locations
\begin{equation} \label{eq:4}
p(x_{t}|y_{t}) \approx \sum_{i=1}^{N} \bar{\omega}^{(i)}_{t} \delta(x_{t}-\tilde{x}^{(i)}_{t}),
\end{equation}
where $\tilde{x}^{(i)}_{t}$ is the peak of the correlation response map corresponding to the $i$-th particle. 

However, the posterior distribution must take into consideration the weights corresponding to the shifted locations, not the original particles. As seen in the top and middle rows of Fig. \ref{fig:2problem}, it is  possible that multiple particles with different correlation maps and weights converge to the same location, which means the posterior distribution would then include multiple particles at a common location but with different weights, which invalidates the assumption that the likelihood depends solely on $x_t^{(i)}$. In other words, the patch centered in the shifted location generates different features and consequently a different weight from the corresponding patch centered in the particle as shown in the middle and bottom rows of Fig. \ref{fig:2problem}. Thus, the posterior distributions estimated by these trackers are not accurate. 

Because the posterior distributions generated by these trackers are not reliable, they resort to resampling at every frame. While resampling is a suitable solution to avoid sample degeneracy that should be performed when necessary, resampling at every frame causes loss of information. It also causes sample impoverishment (i.e., loss of diversity among particles) and may cause all the particles to collapse to a single point within a few frames. In \citep{mozhdehideep4}, we addressed the aforementioned problems by proposing a likelihood particle filter. Although the peaks of the correlation maps in \citep{mozhdehideep4} are used as the proposal and transition distributions, the weights of the peaks are still calculated based on the likelihood of the particles instead of the likelihood of the peaks. Our proposed iterative particle filter is a novel solution for correlation-convolutional trackers to calculate an accurate posterior without performing resampling at every frame.


\begin{figure*}
     \centering    
\centering
\includegraphics[width=0.75\textwidth]{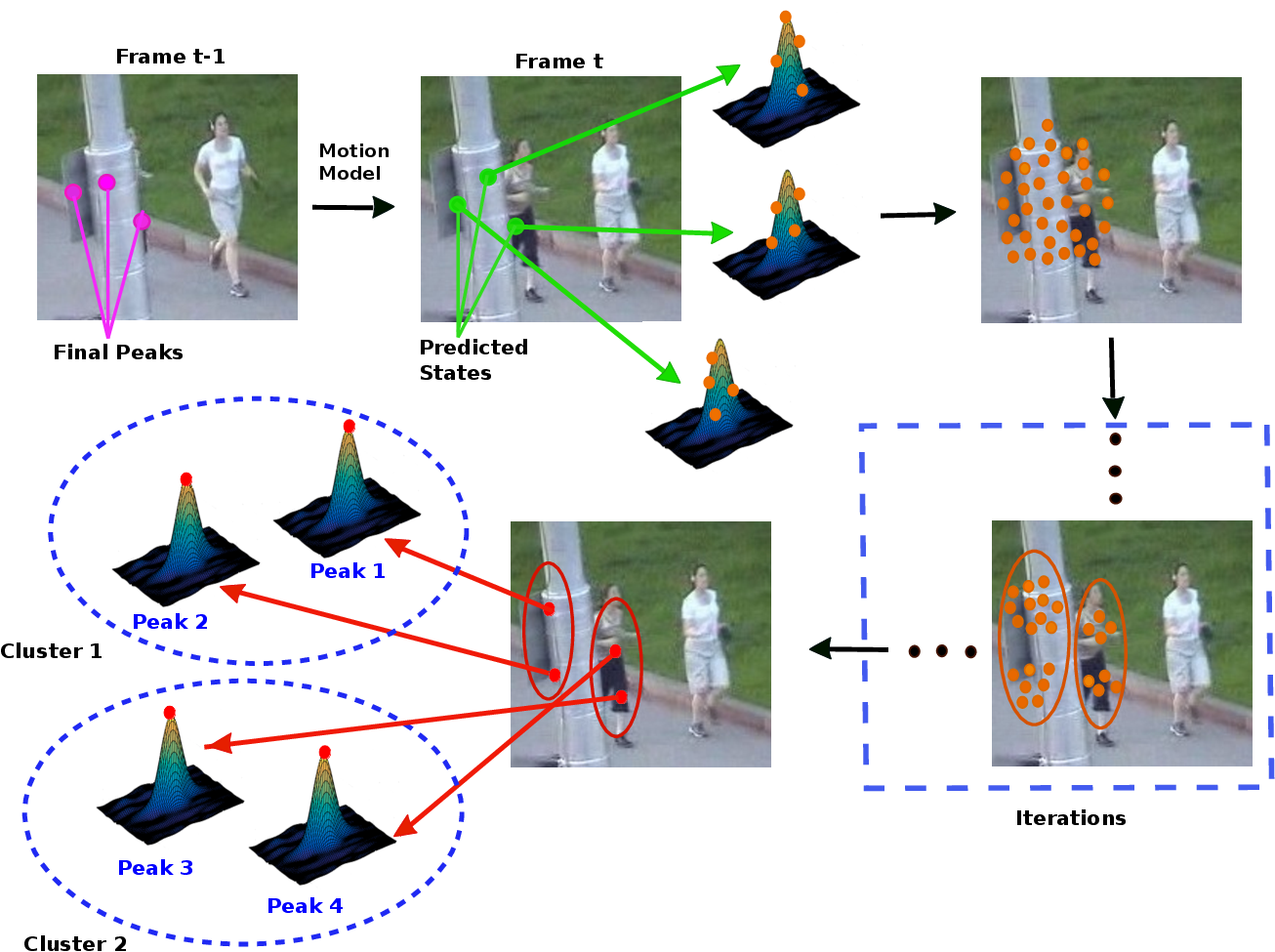}
\begin{small}
\centering
\caption {Illustration of the particle selection process for $J_{t-1} = 3$. The particles are sampled from three distributions whose means are given by the previous correlation map peaks. At time $t$, the particles  converge to four final peaks at the end of the iterations. Two clusters are found by applying K-means to the final particle locations. After selecting the best cluster, the peak of the correlation response map corresponding to the cluster with the highest number of particles is selected as the target state.} 
\label{fig:Clusters}
\end{small}
\end{figure*}

\section{Deep convolutional iterative particle filter}
This section discusses our proposed strategy to generate particles that  better reflect the actual position of the target while avoiding the change of support problem discussed in the previous section. Our approach is based on an iterative particle filter that gradually shifts the particle positions to locations that are closer to the peak of the correlation response map while also updating the response maps so that they become less sensitive to background clutter and better aligned with the target position. As the particles are updated, their corresponding weights are also recomputed based on the new correlation response maps.

\subsection{Iterative particle filter}

\begin{figure*}
     \centering    
\centering
\includegraphics[width=0.8\textwidth]{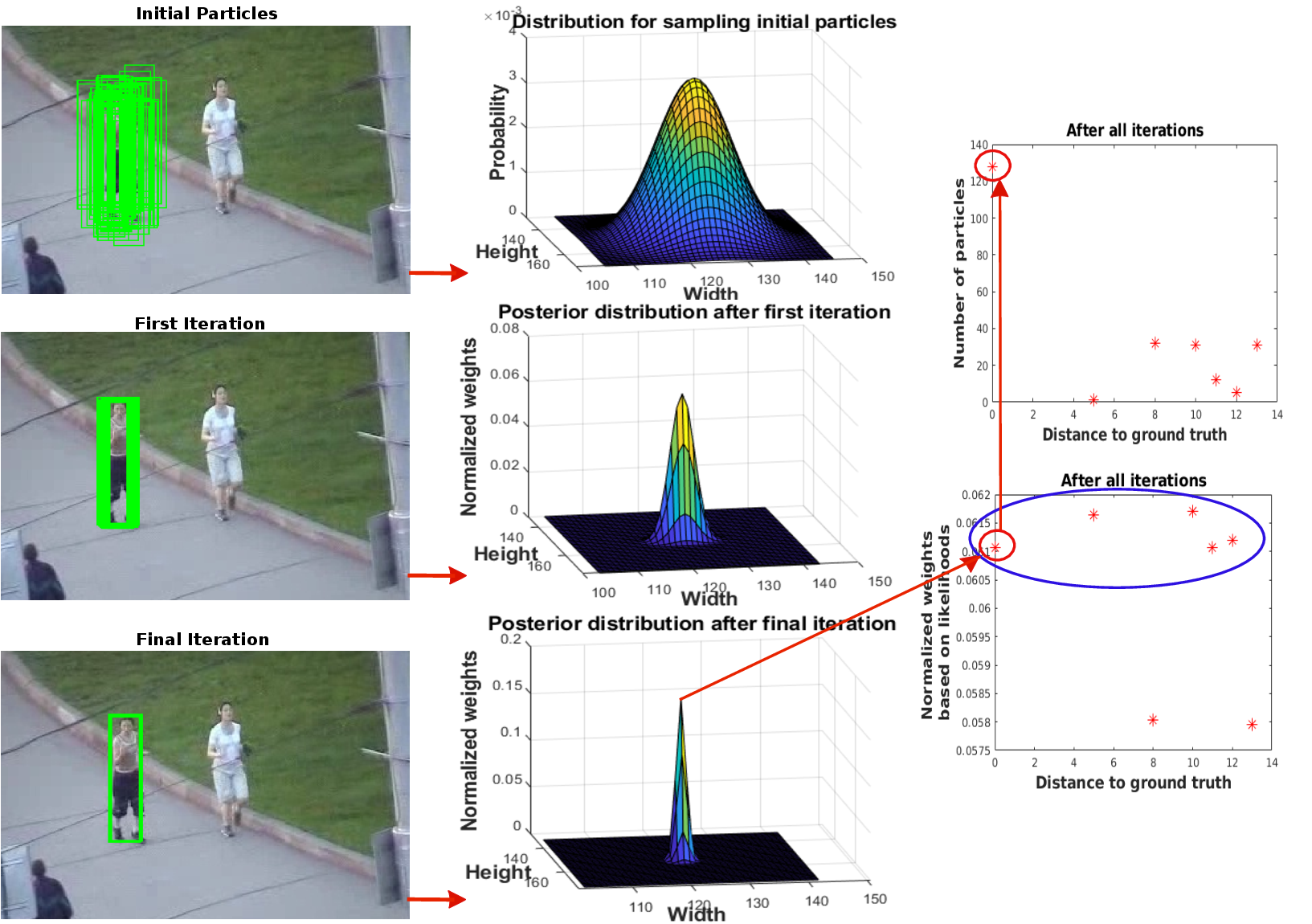}
\begin{small}
\centering
\caption {The images in the left column show how our initial particles converge to the target after the iterations. The plots in the middle column illustrate how the particles reach a sharp final posterior distribution from the wide initial sampling distribution. In the right column, the plots show the normalized weights and the number of particles converging to the peaks after the iterations.
} 
\label{fig:weight2}
\end{small}
\end{figure*}

As Fig. \ref{fig:4iterative} illustrates, correlation filter-based trackers attempt to determine the position of the target by analyzing the displacement between the center of the correlation map, which corresponds to each particle in our tracker, and the peak of the map. A particle filter allows us to generate multiple samples around the predicted target state and hence increase our chances of finding maps with low displacement \citep{mozhdehideep}. If this displacement is sufficiently low, the features extracted from the CNN are similar to the model and the correlation response map is reliable. Our iterative particle filter considerably decreases this displacement and generates more reliable correlation maps for all the particles. As shown in Fig. \ref{fig:4iterative}, after generating the correlation response map for one particle, the distance between the particle position (yellow point) and the peak of the map (red point) is calculated. If the distance is larger than a small threshold $\epsilon$, the correlation response map is not reliable enough to estimate the target position because it was generated based on an image patch centered at a position (yellow point) far from the ground truth location (black point). In such scenarios, the corresponding particle needs to be refined. To that end, the peak of the map is considered the new particle position and its corresponding correlation response map is calculated in a subsequent iteration. Although the peak of the new map (cyan point) is closer to the ground truth, the corresponding particle needs further refinement because the distance between the new particle position (which is now the red point) and the peak of the new map is larger than $\epsilon$. Finally, in the $K_{t}$-th iteration, the calculated distance is smaller than $\epsilon$ and the iterative refinement procedure terminates. The peak of the final map (purple point) is considered the estimated target position for this particle. Since no shifting is performed in the last iteration, the particle support problem discussed above is avoided. Our iterative particle filter is explained in greater detail in the following subsections.

\subsubsection{Particle prediction model}
\label{sec:prediction}

\begin{figure*}
     \centering   
\centering
    \includegraphics[width=0.8\textwidth]{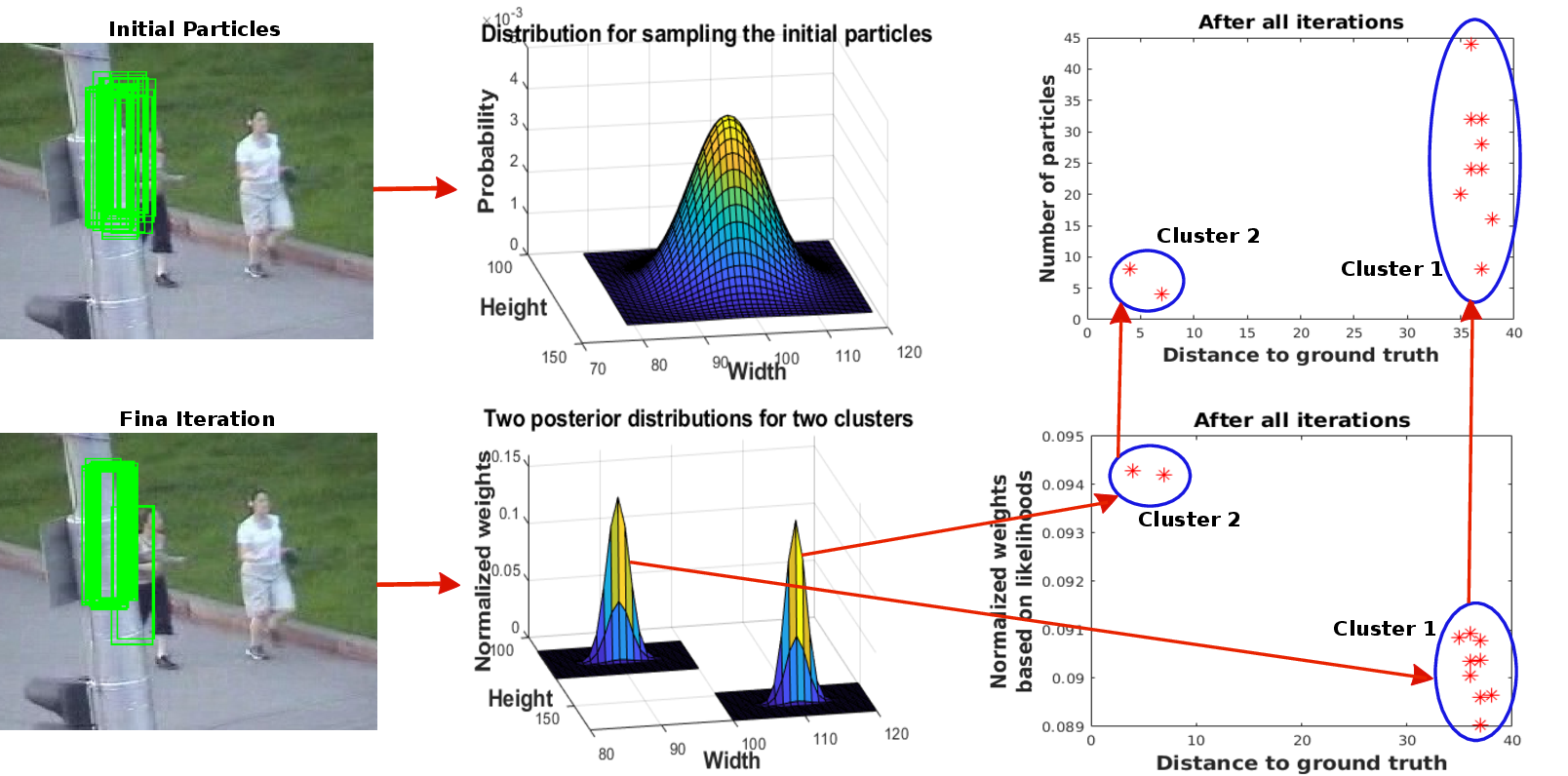}
\begin{small}
\centering
\caption{The images in the left column show that the initial particles reach two clusters after the iterations in a challenging scenario. The plots in the middle column illustrate that the two clusters correspond to two sharp posterior distribution modes from the wide initial sampling distribution after applying K-means to the normalized particle weights. The right column shows that the weights are more reliable for distinguishing the clusters than the number of particles converging to their modes.}
\label{fig:converge3}
\end{small}
\end{figure*}

As shown in the top row of Fig. \ref{fig:Clusters}, the posterior distribution of the target at time $t-1$ is modeled by a mixture of $J_{t-1}$ normal distributions $\mathcal{N}(z_{t-1}^{(j)},\sigma^{2})$ whose means are given by
\begin{equation}\label{eq:5}
    z^{(j)}_{t-1} = \begin{bmatrix}
    x^{(j)}_{t-1},
    \dot{x}^{(j)}_{t-1}
    \end{bmatrix}^T,
\end{equation}
where $j=1,\ldots,J_{t-1}$,  $x^{(j)}_{t-1}=[p^{(j)}_{t-1},s^{(j)}_{t-1}]$ comprises the position $p^{(j)}_{t-1}\in \mathbb{R}^2$ and size $s^{(j)}_{t-1}\in \mathbb{R}^2$ of the target and $\dot{x}^{(j)}_{t-1} \in \mathbb{R}^4$ corresponds to its velocity. We use a constant velocity motion model to predict the means of the $J_{t}$ distributions at the next time instant according to
\begin{equation} \label{eq:6}
  \hat{z}_{t}^{(j)} = A z^{(j)}_{t-1},
\end{equation}
where $A$ is the process matrix defined by 
\begin{equation} \label{eq:7}
   A = \left[ \begin{array}{c|c}
    I_4 & I_4 \\ \hline 
    0_{(4,4)} &  I_4
\end{array}  \right] ,
\end{equation}
where $I_4$ is a $4 \times 4$ identity matrix and $0_{(4,4)}$ is a $4 \times 4$ zero matrix. Our transition distribution is then given by the mixture
\begin{equation} \label{eq:8}
p(x_{t,0}\vert x_{t-1}) = \frac{1}{J_{t-1}}\sum_{j=1}^{J_{t-1}}\mathcal{N}(\hat{z}_{t}^{(j)},\sigma^{2}).
\end{equation}
We then sample $N_{t} \times J_{t-1}$ new particles $x_{t,0}^{(i,j)}\sim p(x_{t,0}\vert x_{t-1})$, where $i=1,\ldots,N_{t}$. 
To increase the efficiency of our strategy, instead of sampling all the particles directly from the mixture distribution, we employ a stratified approach and sample $N_{t}$ particles from each of the $J_{t-1}$ predicted normal distributions. Fig. \ref{fig:Clusters} also illustrates the processes of refining and clustering the particles, which are discussed in more detail in the following sections. Algorithm \ref{alg:1} summarizes our iterative particle filter algorithm. Lines 1-4 correspond to the sampling method described above. The remaining steps of the algorithm are also discussed in the following sections.

\begin{algorithm}
\caption{Deep Convolutional Correlation Iterative Particle (D2CIP).} 
\label{alg:1}
     \begin{algorithmic}[1]
         \Require{Current frame at time $t$, target models, previous final peaks $z^{(j)}_{t-1}$ and their normalized weights $\bar{\omega}_{t-1}^{(j)}$}
         \Ensure{Estimated target state $x^{*}_{t}$, updated target models, current final peaks $z_{t,K_{t}}^{(i,j)}$ and their normalized weights $\bar{\omega}_{t,K_{t}}^{(i,j)}$}
         \State {Find the predicted distributions $\mathcal{N}(\hat{z}_{t}^{(j)},\sigma^{2})$ according to Eqs. \ref{eq:5} to \ref{eq:7}} 
         \For {each predicted state $\hat{z}_t^{(j)}$}
             \State Sample $N_{t}$ initial particles $x_{t,0}^{(i,j)} \sim \mathcal{N}(\hat{z}_{t}^{(j)},\sigma^{2})$
         \EndFor 
         \For {each initial particle $x_{t,0}^{(i,j)}$}
             \State Find its corresponding final peak using Alg. \ref{alg:2} 
         \EndFor
         \For {each final peak $x_{t,K_{t}}^{(i,j)}$}
             \State Calculate the peak weight $\omega_{t,K_{t}}^{(i,j)}$ 
         \EndFor
         \State {Estimate the target state $x^{*}_{t}$ based on the final peaks $x_{t,K_{t}}^{(i,j)}$ using Alg. \ref{alg:3}}
         \State {Find the updated target models based on the final peaks $x_{t,K_{t}}^{(i,j)}$}
         \State {Resample if the effective sample size is lower than $\gamma$}
     \end{algorithmic}
\end{algorithm}

\subsubsection{Iterative particle refinement}
\label{sec:refinement}

Particle $x_{t,0}^{(i,j)}$ is used to sample a patch from the current frame at time $t$ and to generate the corresponding convolutional features. These features are compared with the target models to calculate the correlation response map $R^{(i,j)}_{t,0}$ using the correlation filter proposed in \citep{ECO}. We maintain one target model for each of the predicted distributions generated using Eq. \ref{eq:6}, but to simplify the notation in this section, we refrain from explicitly differentiating the models. Let $p(y_{t}|x_{t,0}^{(i,j)})$ be the likelihood of $x_{t,0}^{(i,j)}$, which is given by the sum of the elements of $R^{(i,j)}_{t,0}$. We discard the samples for which $p(y_{t}|x_{t,0}^{(i,j)})<L_{min}$, where $L_{min}$ is the threshold to consider a correlation response map acceptable. As illustrated in Fig. \ref{fig:4iterative}, at each iteration $k=1,\ldots,K_t$, the remaining samples are shifted to $x_{t,k}^{(i,j)}$, which is defined  as 
\begin{equation} \label{eq:9}
x_{t,k}^{(i,j)}=[p_{t,k}^{(i,j)},s_{t,0}^{(i,j)}],
\end{equation} 
where $p_{t,k}^{(i,j)} = \argmax(R_{t,k-1}^{(i,j)})$, i.e.,  the peak of the associated correlation response map at step $k-1$ of the iterative refinement process. We then have $d_{t,k}^{(i,j)}=||p_{t,k}^{(i,j)}-p_{t,k-1}^{(i,j)}||$ as the Euclidean distance between $p_{t,k}^{(i,j)}$ and $p_{t,k-1}^{(i,j)}$. For each particle, the refinement procedure continues until $d_{t,k}^{(i,j)}<\epsilon$. Since particles in close proximity tend to generate correlation response maps whose peaks share a common location, all the particles converge to a small number high-likelihood positions. These peaks determine the means of the normal distributions used to generate the prediction model described in Section \ref{sec:prediction}. 

As seen in Fig. \ref{fig:Clusters}, let $\mathcal{N}(x_{t,k}^{(i,j)},\sigma^{2})$ be the normal distributions after the convergence of the $k$-th iteration. We select the mean of these normal distributions as the particles for the next iteration if $d_{t,k}^{(i,j)}\geq \epsilon$. After the iterations, the particles reach the final peaks $x_{t,K_{t}}^{(i,j)}=[p_{t,K_{t}}^{(i,j)},s_{t,0}^{(i,j)}]$, which do not need further refinement because $d_{t,K_{t}}^{(i,j)} < \epsilon$. Thus, we have 
\begin{equation}\label{eq:10}
p_{t,K_{t}}^{(i,j)}=p_{t,0}^{(i,j)}+\sum_{k=1}^{K_{t}}d_{t,k}^{(i,j)}.
\end{equation}

Algorithm \ref{alg:2} explains how to reach the final peaks in our iterative particle filter. Additionally, all the normal distributions $\mathcal{N}(x_{t,K_{t}}^{(i,j)},\sigma^{2})$ are used in the process of updating the target models as well. Our baseline tracker examines only the estimated target state to update the target models, while our iterative particle filter provides all $\mathcal{N}(x_{t,K_{t}}^{(i,j)},\sigma^{2})$ for the tracker to examine in the target model update process.

\subsubsection{Weight update model}

The posterior distribution for the particle filter is approximated by
\begin{equation} \label{eq:11}
p(x_{t}|y_{t}) \approx \sum \bar{\omega}^{(i,j)}_{t,K_{t}} \delta(x_{t}-x^{(i,j)}_{t,K_{t}}),
\end{equation}
where $\bar{\omega}^{(i,j)}_{t,K_{t}}$ represents the normalized weights of the final correlation map peaks. Unlike previous convolution-correlation particle filters, since our approach keeps track of the particles that converge to a common location, it allows us to update the particle posterior distribution based on the likelihood of their final locations and their corresponding prior weights, i.e., 
\begin{equation} \label{eq:12}
     \omega^{(i,j)}_{t,K_{t}} \propto p(y_{t}|x^{(i,j)}_{t,K_{t}})\max_{\bar{\omega}_{t-1}^{(j)} \in \mathcal{X}_{t-1}^j}{\bar{\omega}_{t-1}^{(j)}},
\end{equation}
where $ p(y_{t}|x^{(i,j)}_{t,K_{t}})$ is the likelihood of the final peak based on its correlation response map and $\mathcal{X}_{t-1}^j$ is the set of weights of $\mathcal{N}(z_{t-1}^{(j)},\sigma^{2})$ that converge to $x^{(i,j)}_{t,K_{t}}$. This approach allows us to refrain from unnecessarily resampling the particles at every frame. Instead, we perform resampling only when the effective sample size is lower than a threshold $\gamma$ as in \citep{mozhdehideep}.

\begin{algorithm}
\caption{Iterative Particle Refinement.} 
\label{alg:2}
     \begin{algorithmic}[1]
         \Require{Current frame $t$, initial particles $x_{t,0}^{(i,j)}$, target models}
         \Ensure{Final current peaks $x_{t,K_{t}}^{(i,j)}$} 
             \For {each particle $x_{t,0}^{(i,j)}$} 
                 \State $d_{t,k}^{(i,j)} = \infty$
                 \While {$d_{t,k}^{(i,j)} > \epsilon$}
                     \State Generate the correlation response map $R_{t,k-1}^{(i,j)}$
                     \State Calculate the likelihood $p(y_{t}|x_{t,k-1}^{(i,j)})$ based on $R_{t,k-1}^{(i,j)}$
                     \If {$p(y_{t}|x_{t,k}^{(i,j)}) > L_{min}$}
                         \State $p_{t,k}^{(i,j)} = \argmax(R_{t,k-1}^{(i,j)})$
                         \State $d_{t,k}^{(i,j)} = ||p_{t,k}^{(i,j)} - p_{t,k-1}^{(i,j)}||$
                     \Else
                        \State Discard particle $x_{t,k-1}^{(i,j)}$
                     \EndIf
                 \EndWhile
                 \State {Find $J_{t,K}$ for each final peak}
             \EndFor
\end{algorithmic}
\end{algorithm}

\subsection{Target state estimation}
\label{sec:Clusters2}

Using Eq. \ref{eq:12} in simple frames that do not involve any challenging scenario results in particle weights very similar to one another. Again, this is because the particles converge to a few nearby peaks after the iterations. Hence, the correlation maps related to these final peaks are similar as well. Thus, evaluation of the particles based on likelihoods calculated from the correlation maps is not sufficiently accurate in such simple frames. However, after the iterations, it is possible to determine the location to where most particles converge. As Fig. \ref{fig:weight2} illustrates, our initial particles gradually converge to a few peaks at the end of the iterative refinement procedure. The plots in the middle column of the figure show how the iterations decrease the area covered by the particles. As the plots indicate, after the iterations, the particles reach a sharp posterior distribution from a wide initial distribution. The plots in the left column of the figure show that the weights based on the correlation maps are similar to one another after the iterations. As seen in the bottom right plot, which shows the weights based on Eq. \ref{eq:12}, the weight of the peak located exactly on the ground truth is slightly lower than the weights of farther peaks (shown within the blue ellipse). However, most particles converge to the peak closest to the ground truth location as shown in the plot at the top of the right column. Thus, the final state $x_{t}^*$ is calculated by \begin{equation} \label{eq:likelihoodweight}
     x_{t}^* = \argmax_{x_{t,K_{t}}^{(i,j)}} J_{t,K_{t}},
\end{equation}
where $J_{t,K_{t}}$ is the number of particles $x_{t,0}^{(i,j)}$ that converge to the common final peak $x_{t,K_{t}}^{(i,j)}$.

However, when the tracker faces a challenging scenario, the area covered by the particles does not necessarily decrease after the iterations. This is because the particles may converge to different image regions. In the challenging scenario illustrated in Fig. \ref{fig:converge3}, the particles converge to the pole and the jogger, which correspond to two clusters of particles. In such scenarios, we first determine the number of clusters and select the one that best represents the posterior. Since the particles converge to distinct image regions, our proposed method can partition them using K-means clustering \citep{Bishop}. We determine the number of modes in the distribution using simplified silhouette analysis based on the Euclidean distances among particles \citep{wang2017analysis}. The plots in the middle column of Fig. \ref{fig:converge3} illustrate how the particles form a posterior distribution with two sharp modes from the wide initial sampling distribution. This posterior distribution is calculated based on the particle weights according to Eq. \ref{eq:12}. As seen in the top plot of the right column of Fig. \ref{fig:converge3}, the number of particles converging to each cluster is not sufficiently accurate to find the image region corresponding to the target. As the plot indicates, only a few particles converge to the region surrounding the jogger. The bottom plot of the right column of the figure shows that the weights calculated by Eq. \ref{eq:12}, on the other hand, provide an accurate method to distinguish the clusters. Because of the considerable distance between the clusters, the correlation response maps within different clusters are not similar to each other. Thus, particle evaluation based on the likelihoods according to Eq. \ref{eq:12} is reliable because correlation maps closer to the target generate higher likelihoods. Thus, we first find the clusters using K-means after performing the iterations, and the mode of each cluster is then selected based on the number of particles reaching the final peaks. The best cluster is then selected based on the correlation response maps corresponding to each mode according to Eq. \ref{eq:12}. Algorithm \ref{alg:3} summarizes our mode clustering and state estimation method.

\begin{algorithm}[b]
\caption{Target State Estimation.} 
\label{alg:3}
     \begin{algorithmic}[1]
         \Require{Final peaks $x_{t,K_{t}}^{(i,j)}$ at time $t$, their weights $\omega_{t,K_{t}}^{(i,j)}$ and $J_{t,K_{t}}$}
         \Ensure{Current target state $x_{t}^*$} 
                \State {Apply K-means to all final current peaks $x_{t,K_{t}}^{(i,j)}$ to find the clusters}
                \State {Find the mode of each cluster based on $J_{t,K_{t}}$}
                \State {Compare the weight of the calculated modes based on $\omega_{t,K_{t}}^{(i,j)}$ to select the best mode}
                \State {Consider the best mode as the current target state $x^{*}_{t}$}
\end{algorithmic}
\end{algorithm}

\begin{figure*}[ht]
     \centering 
\begin{minipage}{1\linewidth}
\centering
\includegraphics[trim=30 2 30 2, clip, width=.27\textwidth]{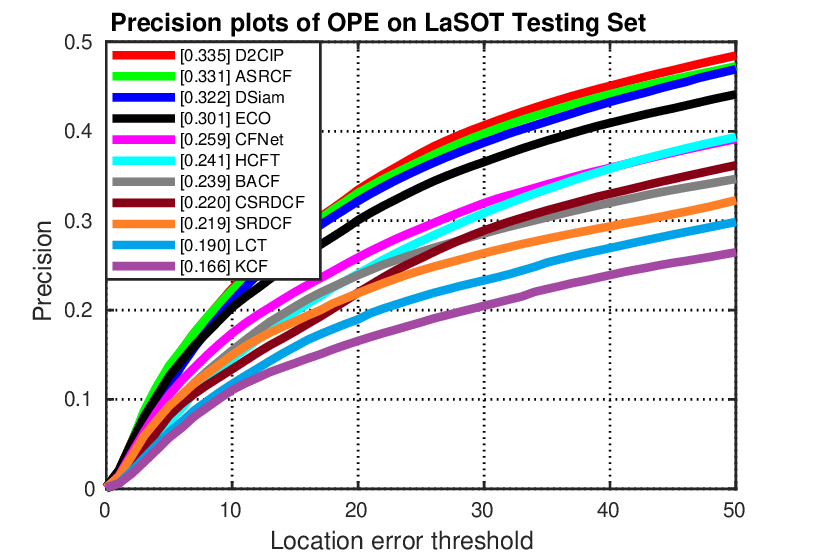}
\includegraphics[trim=30 2 30 2, clip, width=.27\textwidth]{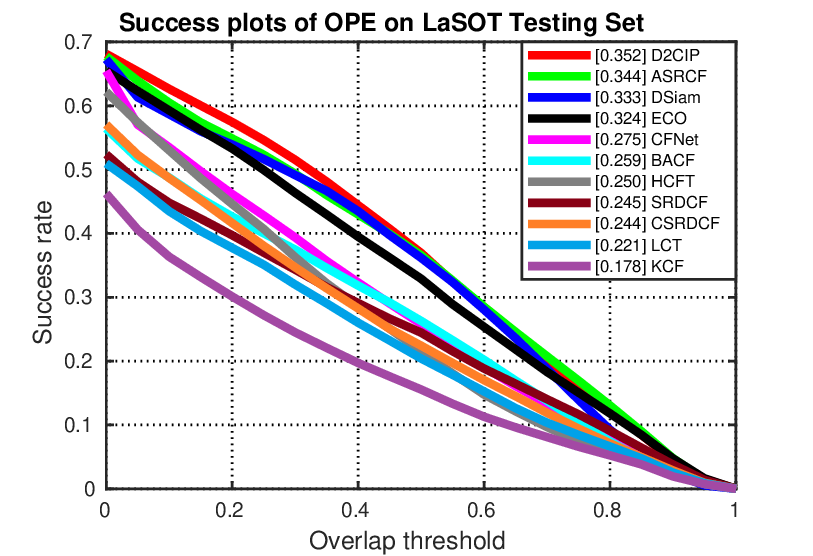}
\includegraphics[trim=30 2 30 2, clip, width=.27\textwidth]{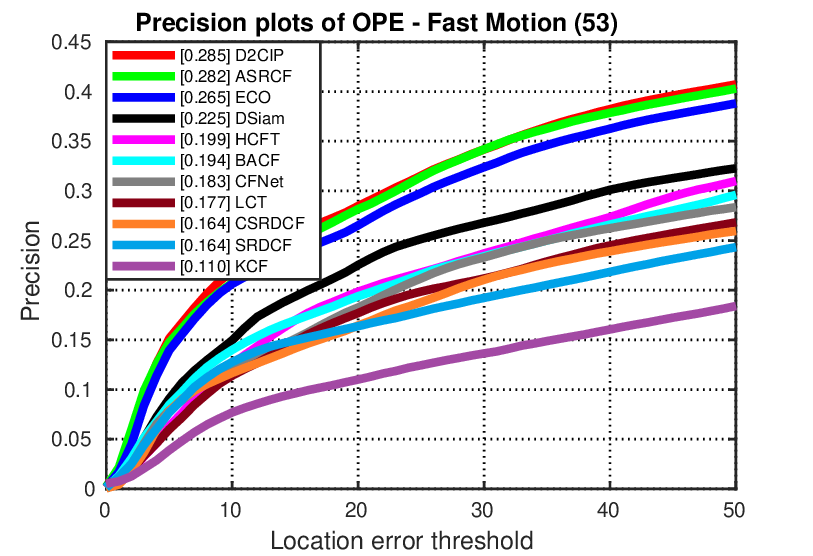}
\end{minipage}  
\begin{minipage}{1\linewidth}
\centering
\includegraphics[trim=30 2 30 2, clip, width=.27\textwidth]{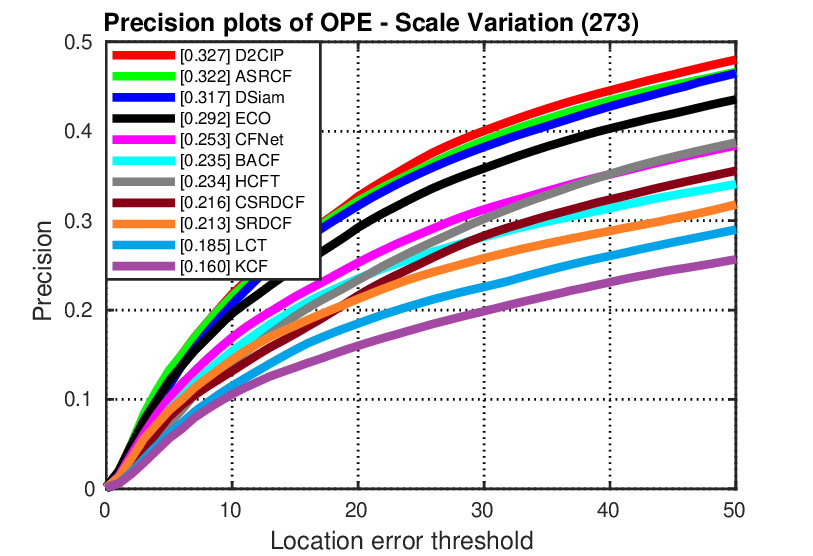}
\includegraphics[trim=30 2 30 2, clip, width=.27\textwidth]{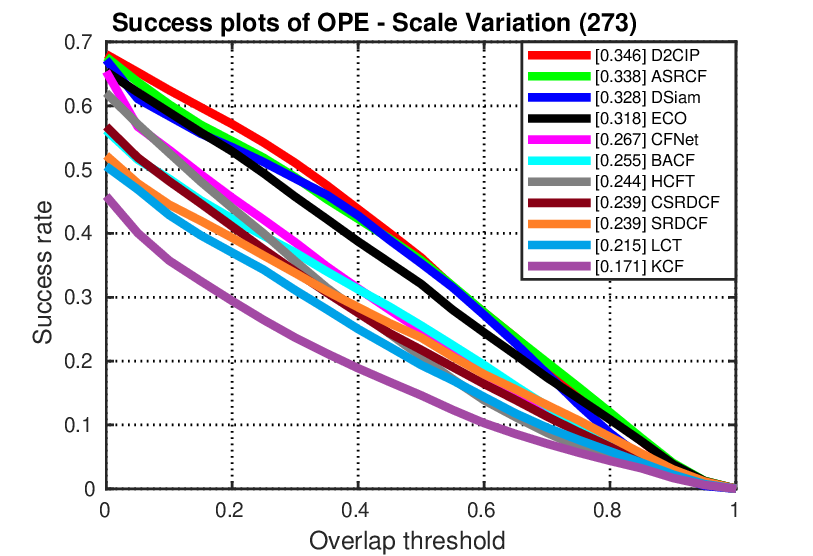}
\includegraphics[trim=30 2 30 2, clip, width=.27\textwidth]{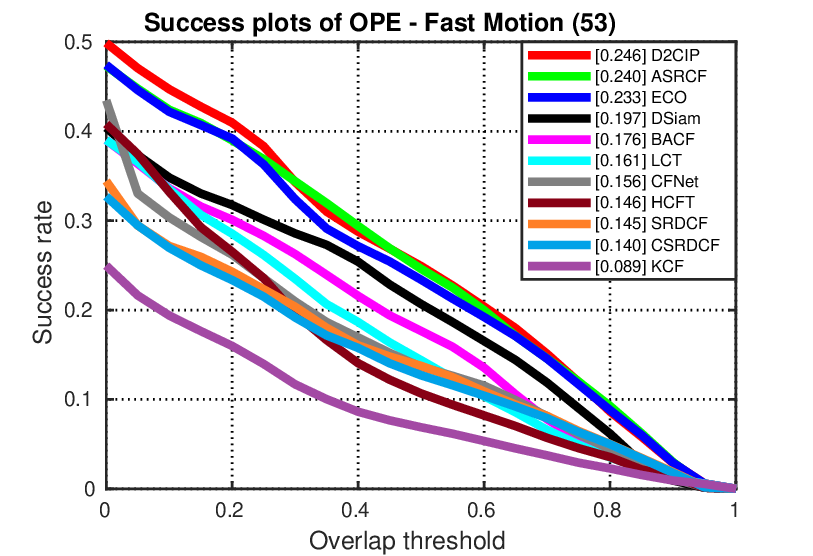}
\end{minipage} 
\begin{minipage}{1\linewidth}
\centering
\includegraphics[trim=30 2 30 2, clip, width=.27\textwidth]{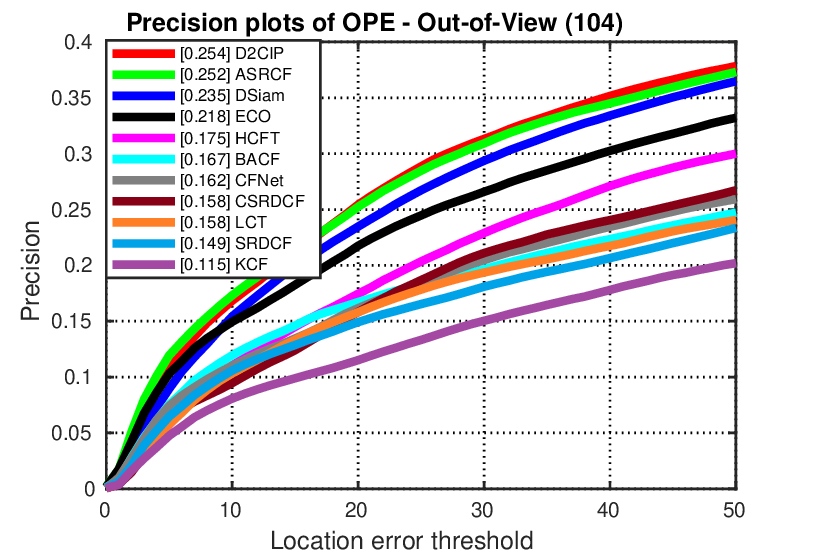}
\includegraphics[trim=30 2 30 2, clip, width=.27\textwidth]{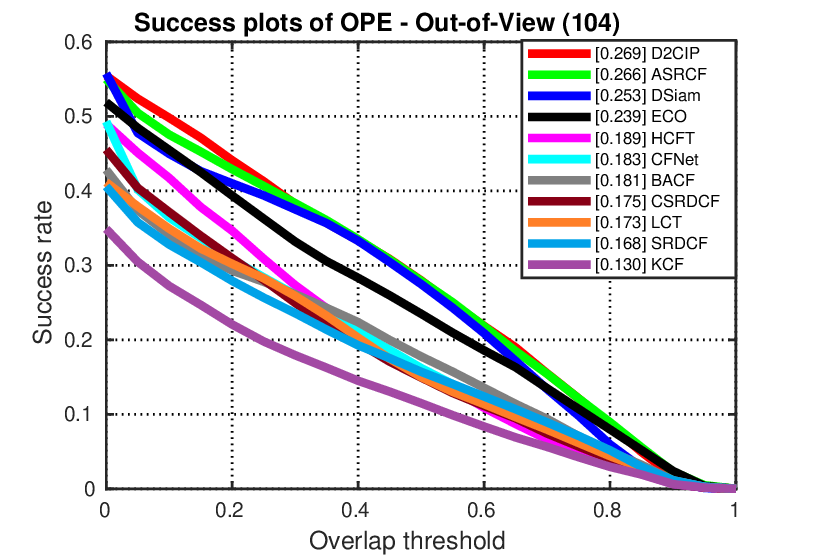}
\includegraphics[trim=30 2 30 2, clip, width=.27\textwidth]{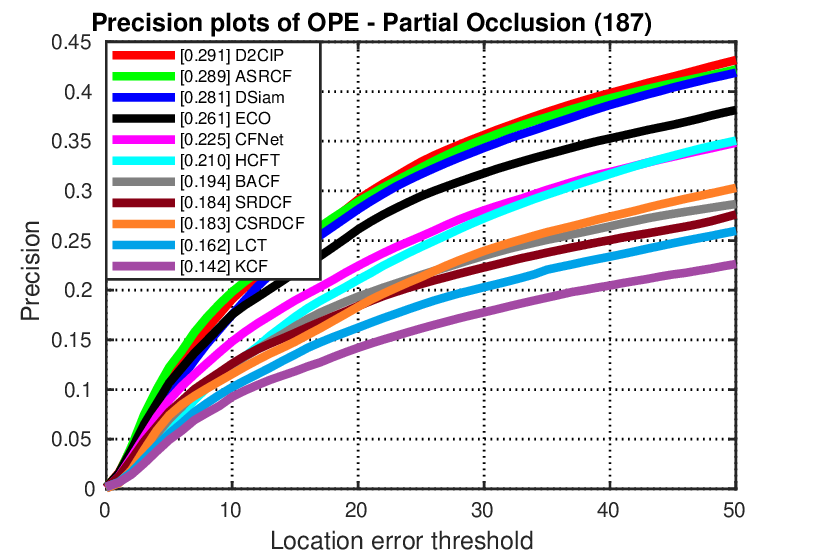}
\end{minipage}   
\begin{minipage}{1\linewidth}
\centering
\includegraphics[trim=30 2 30 2, clip, width=.27\textwidth]{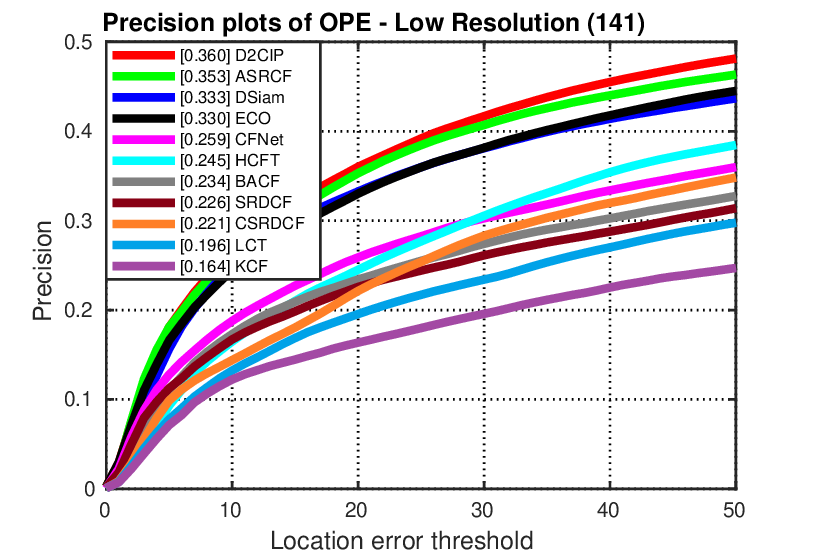}
\includegraphics[trim=30 2 30 2, clip, width=.27\textwidth]{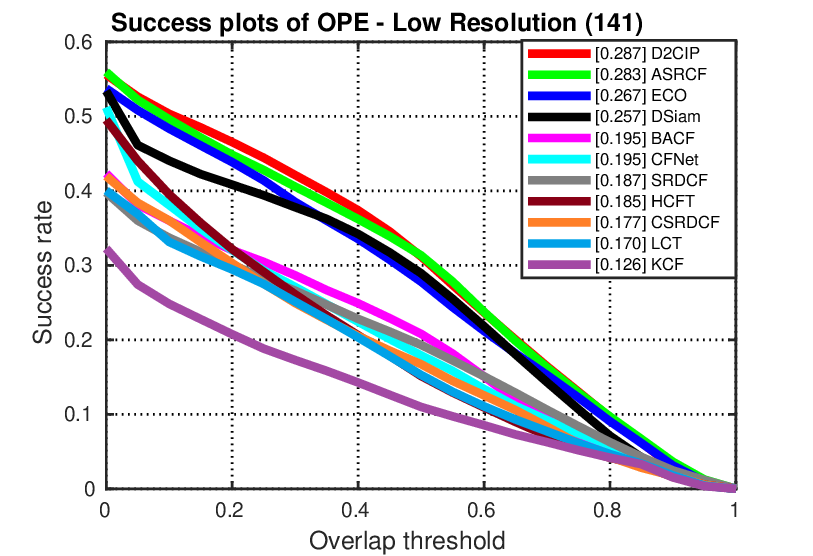}
\includegraphics[trim=30 2 30 2, clip, width=.27\textwidth]{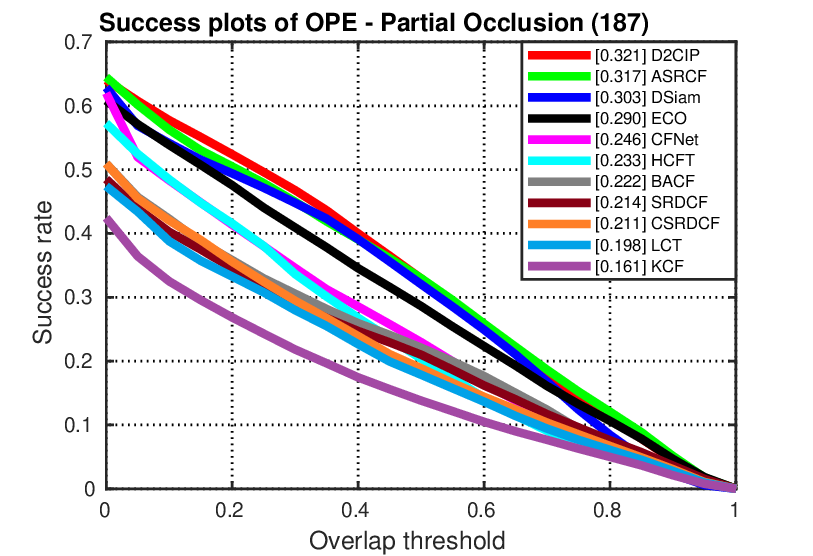}
\begin{small}
\caption{Quantitative assessment of the performance of our tracker in comparison with state-of-the-art trackers using a one-pass evaluation on the LaSOT benchmark dataset.}
\label{fig:OPE3}
\end{small}
\end{minipage}   
\end{figure*}

\section{Results and Discussion}

We evaluate our algorithm on three publicly available visual tracking benchmarks: the large-scale single object tracking benchmark (LaSOT) \citep{LaSOT}, the recently published TREK-150 dataset \citep{TREK150}, and the visual tracker benchmark v1.1 beta (OTB100) \citep{OTB}. All the results shown in this section correspond to a particle filter with $200$ particles.  


\subsection{LaSOT evaluation}

LaSOT is currently the largest publicly available benchmark for object tracking. It provides high-quality dense manual annotations with $14$ attributes representing challenging aspects of tracking. The benchmark consists of $1,400$ videos with an average of $2,512$ frames per sequence. The benchmark provides two different metrics to evaluate visual trackers: precision and success. For more details on the evaluation metrics, we refer the reader to \citep{LaSOT}. 
Fig. \ref{fig:OPE3} presents a quantitative assessment of our proposed approach using a one-pass evaluation (OPE) on LaSOT in comparison with $10$ state-of-the-art trackers including ECO, ASRCF \citep{ASRCF2019CVPR}, DSiam \citep{DSiam2017ICCV}, CFNet \citep{CFNet2017CVPR}, HCFT \citep{HCFT}, BACF \citep{BACF}, CSRDCF \citep{CSRDCF}, SRDCF \citep{SRDCF}, LCT \citep{conf/cvpr/MaYZY15} and KCF \citep{KCF}. In the one-pass evaluation, the tracker is initialized with the ground truth location of the target at the first frame of the image sequence and allowed to keep track of the target over the remaining frames without reinitialization. As seen in Fig. \ref{fig:OPE3}, our tracker outperforms all the other trackers in terms of overall precision and success. In particular, it outperforms ASRCF by $1.2\%$ and $2.3\%$, respectively. Similar to our proposed tracker, ASRCF is a recent state-of-the-art correlation-convolutional visual tracker that uses ECO as a baseline method. Our most significant improvements in comparison with ASRCF occur in low resolution and scale variation scenarios, which show improvements of $2\%$ and $1.6\%$ in precision and $1.4\%$ and $2.4\%$ in success, respectively. In comparison with our baseline tracker, our precision improvement reaches $9.1\%$, $12.0\%$, and $11.5\%$ in low resolution, scale variation, and partial occlusion scenarios, respectively. In terms of the success metric, our improvement in such scenarios reaches $7.5\%$, $8.8\%$, and $10.7\%$ in comparison with ECO.

\begin{figure*}[ht]
     \centering 
\begin{minipage}{1\linewidth}
\centering
\includegraphics[trim=30 2 30 2, clip, width=.27\textwidth]{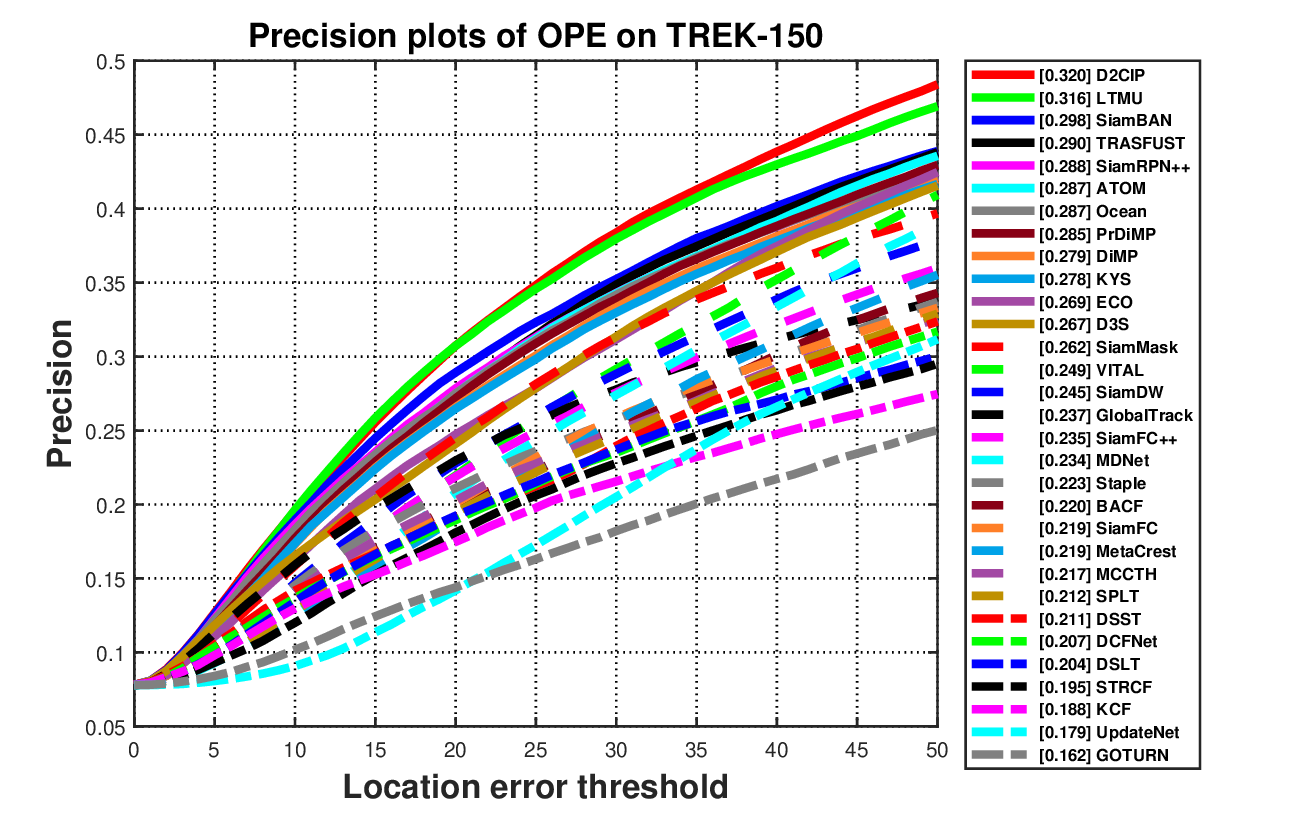}
\includegraphics[trim=30 2 30 2, clip, width=.27\textwidth]{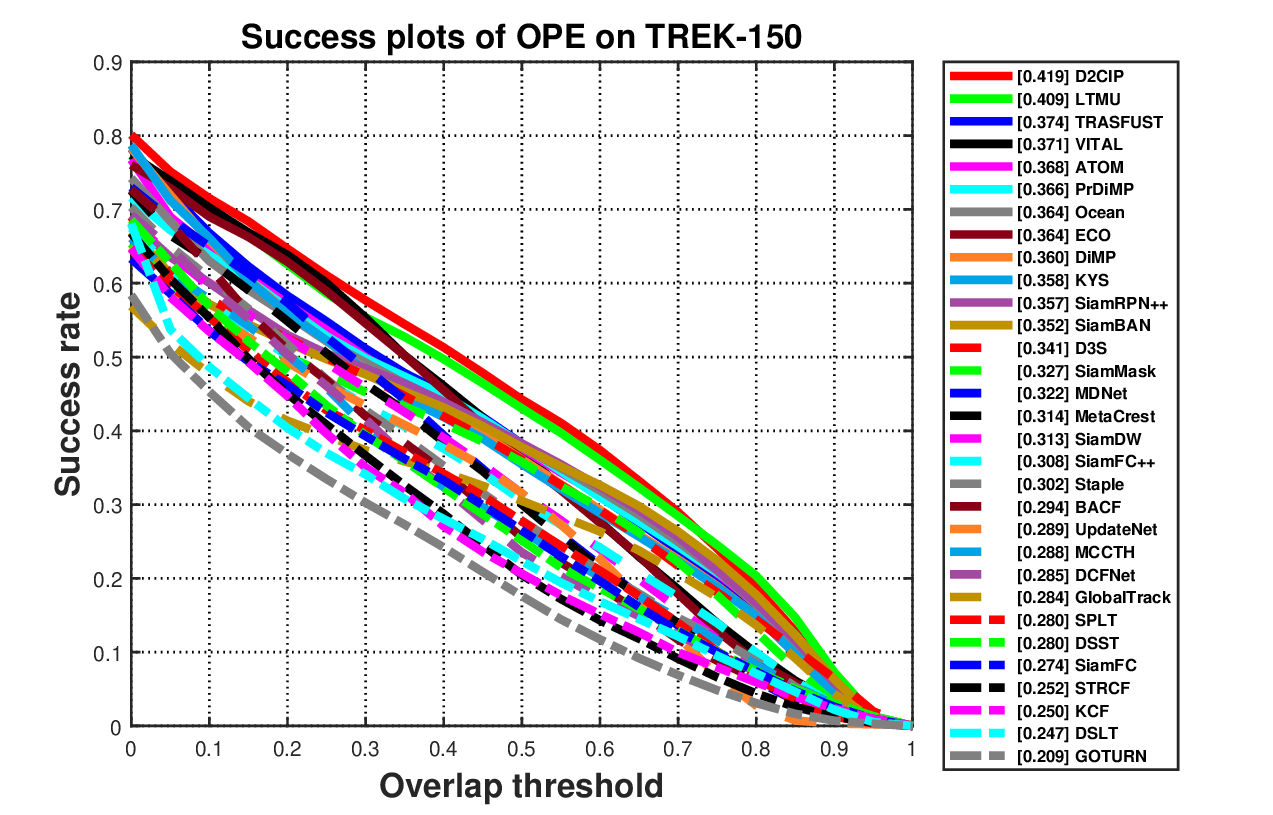}
\includegraphics[trim=30 2 30 2, clip, width=.27\textwidth]{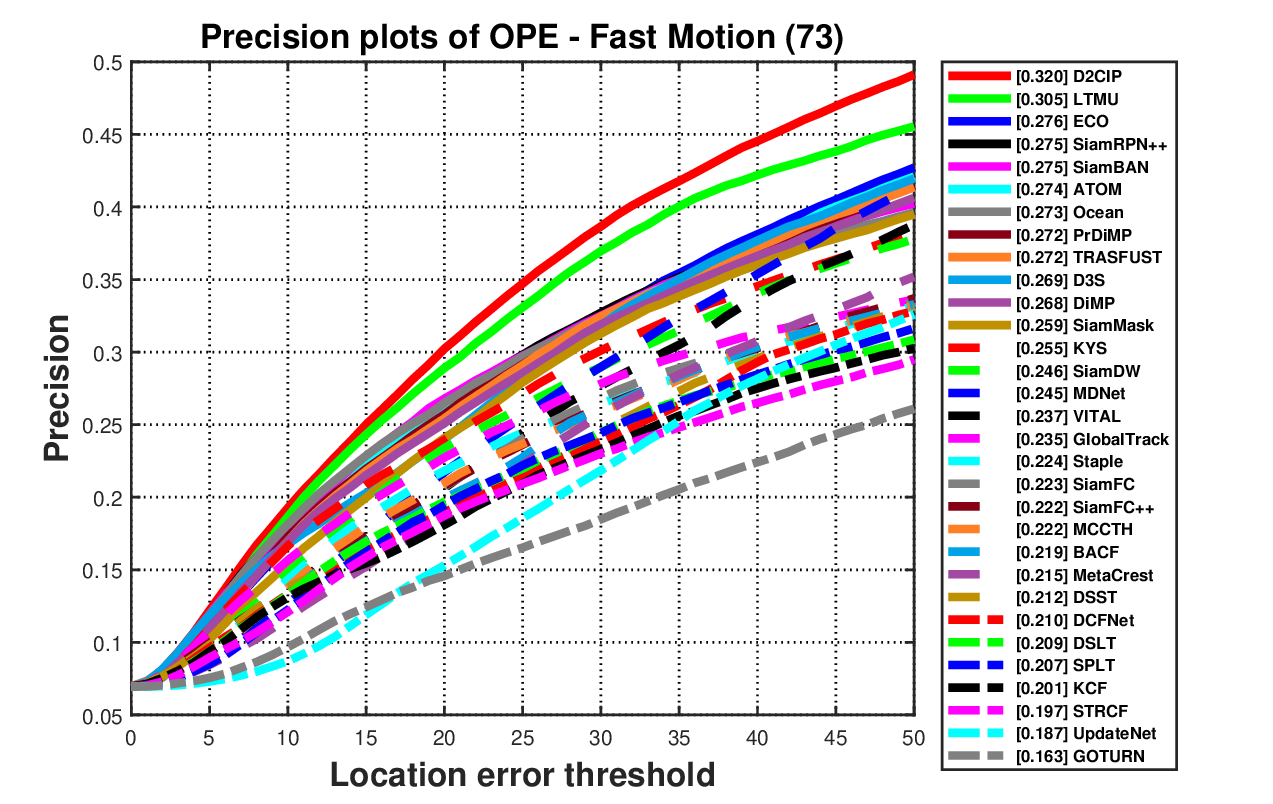}
\end{minipage}  
\begin{minipage}{1\linewidth}
\centering
\includegraphics[trim=30 2 30 2, clip, width=.27\textwidth]{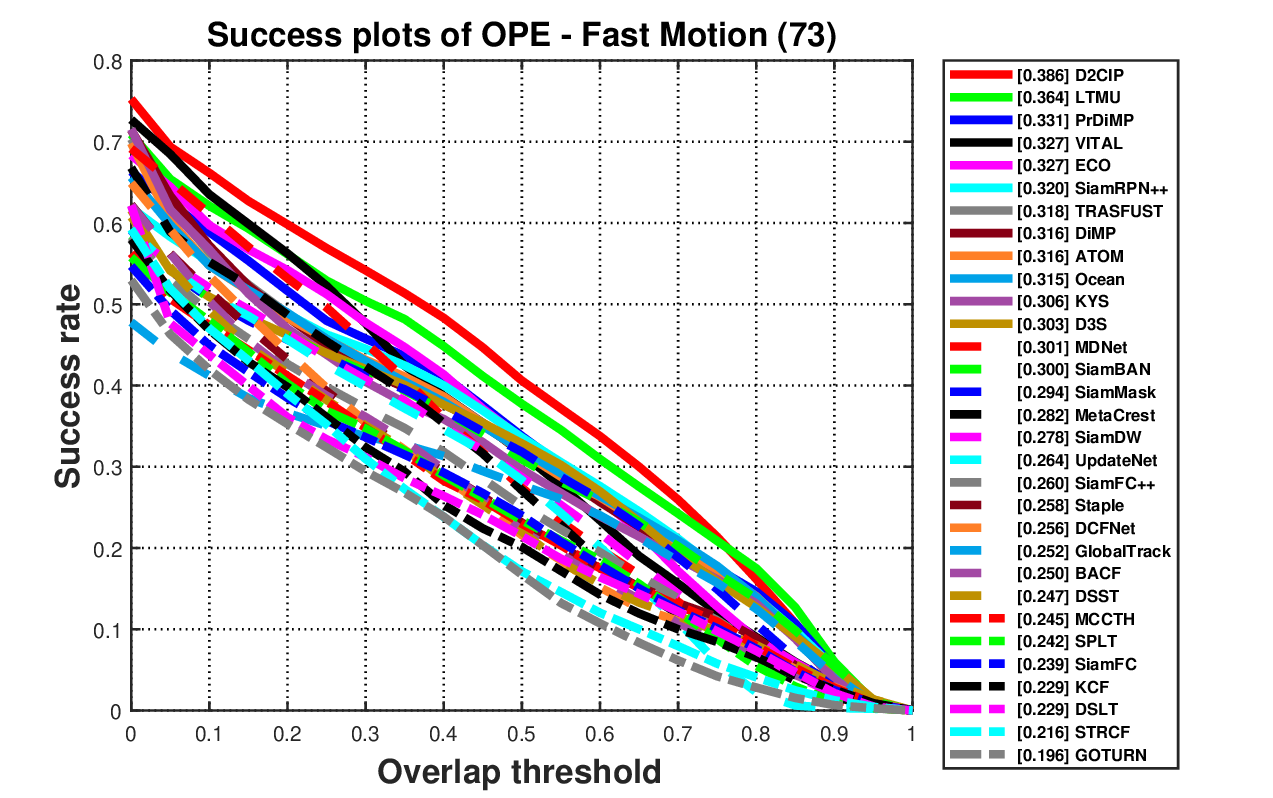}
\includegraphics[trim=30 2 30 2, clip, width=.27\textwidth]{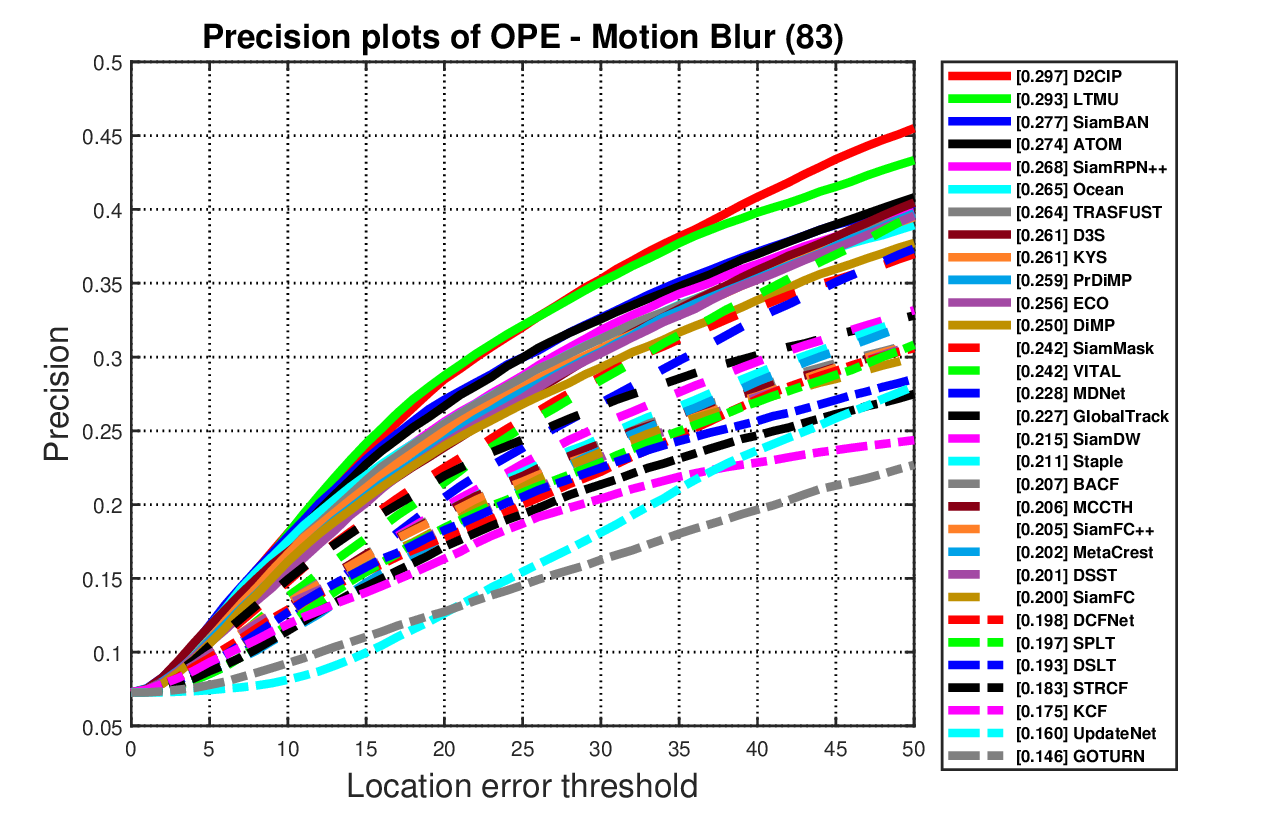}
\includegraphics[trim=30 2 30 2, clip, width=.25\textwidth]{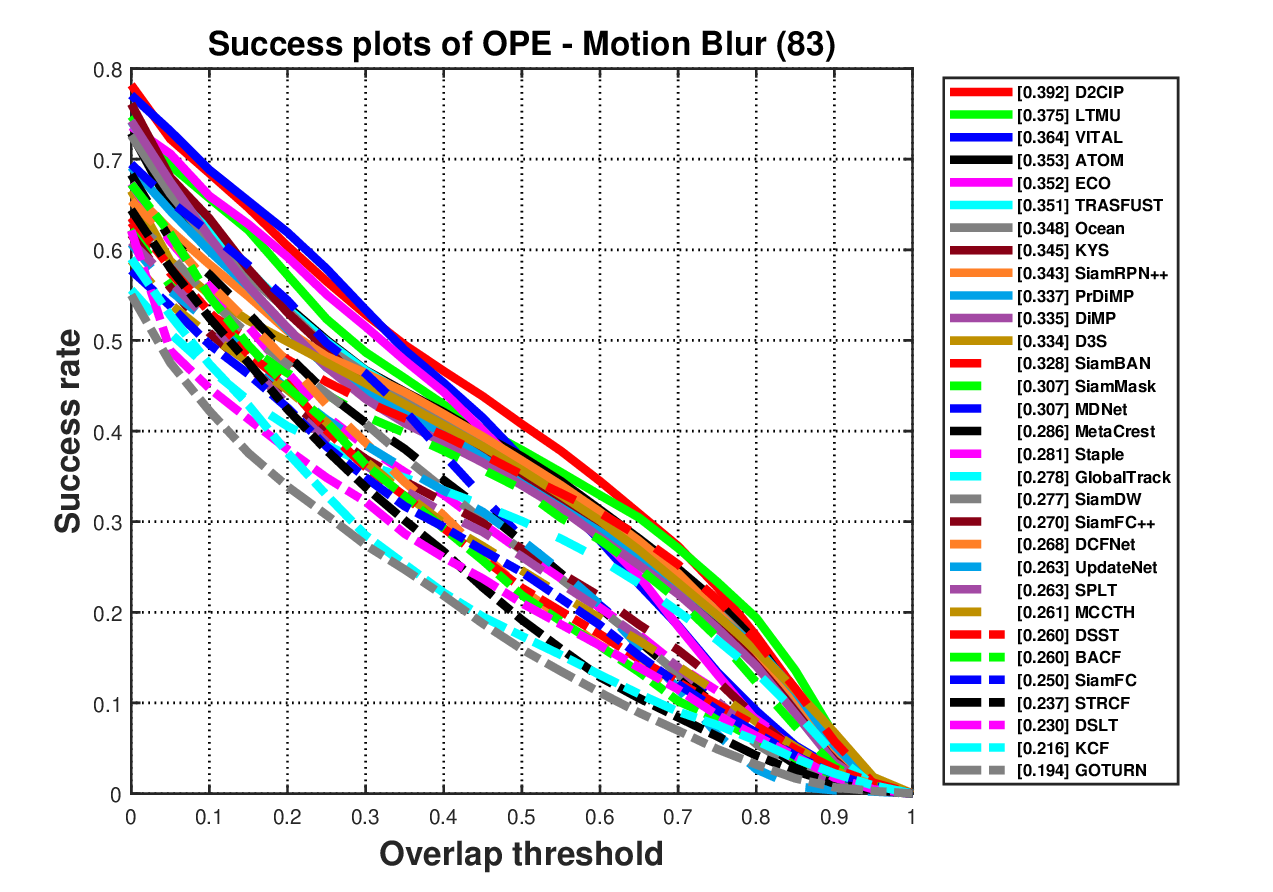}
\end{minipage} 
\begin{minipage}{1\linewidth}
\centering
\includegraphics[trim=30 2 30 2, clip, width=.27\textwidth]{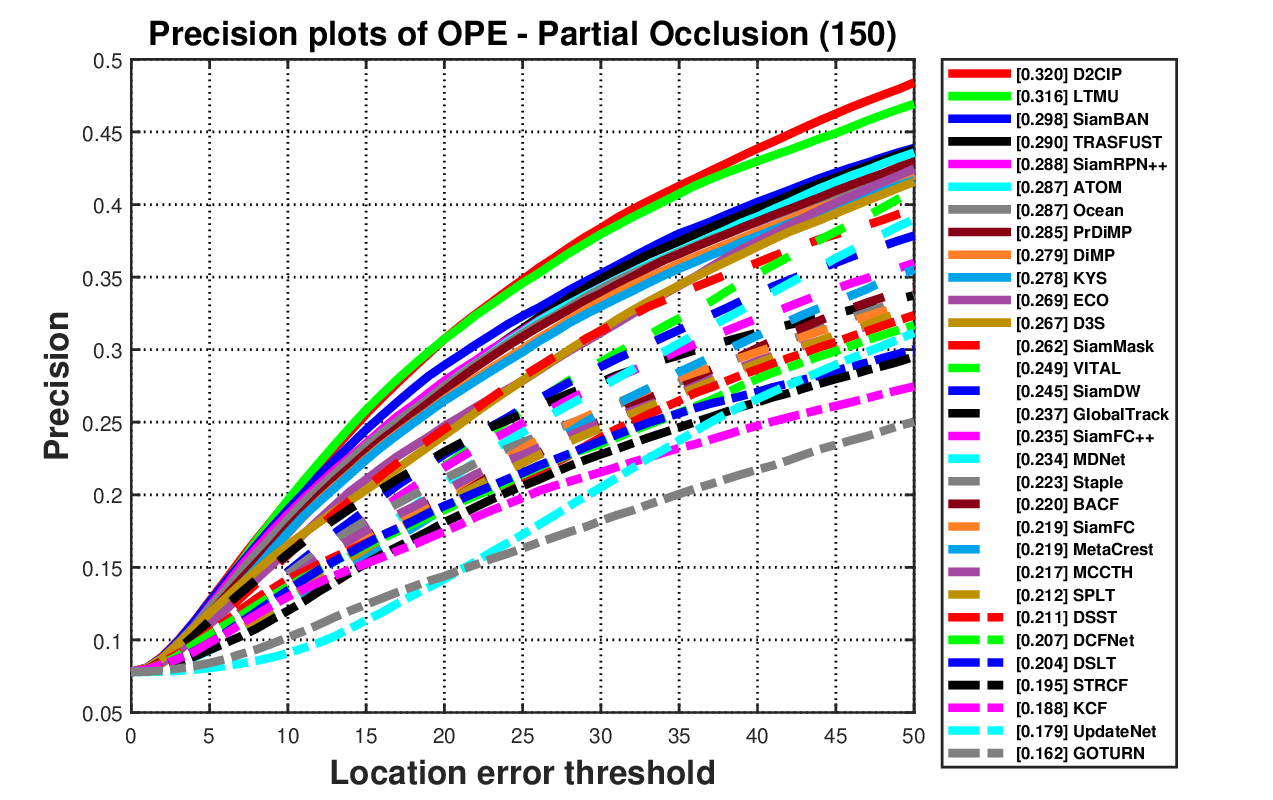}
\includegraphics[trim=30 2 30 2, clip, width=.27\textwidth]{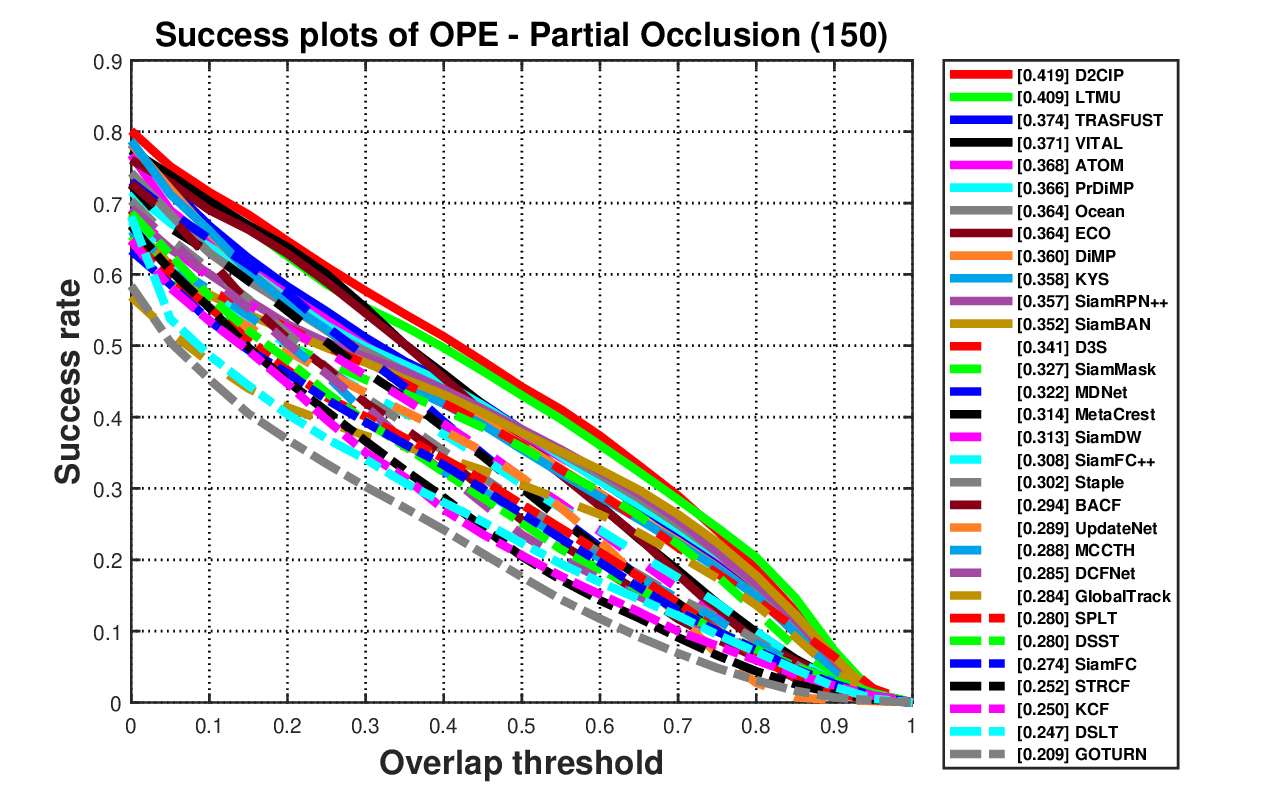}
\includegraphics[trim=30 2 30 2, clip, width=.27\textwidth]{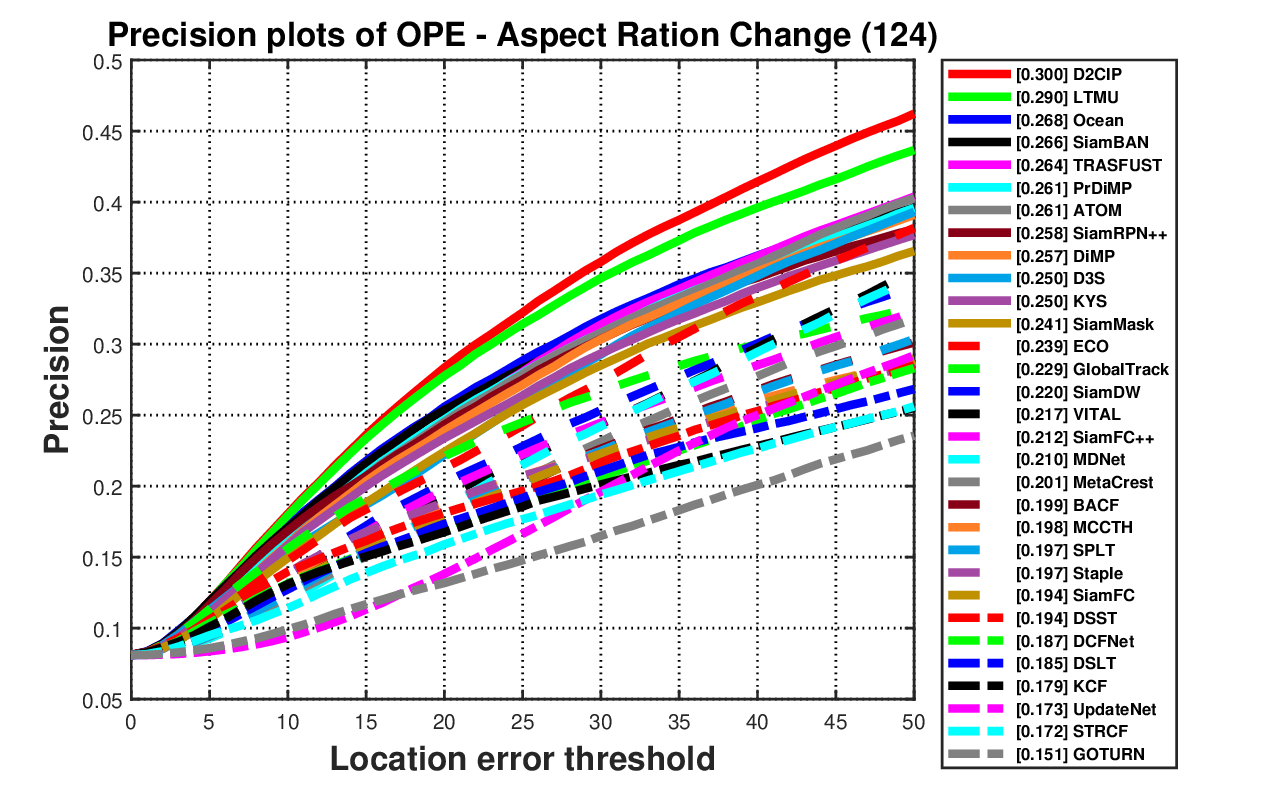}
\end{minipage}   
\begin{minipage}{1\linewidth}
\centering
\includegraphics[trim=30 2 30 2, clip, width=.27\textwidth]{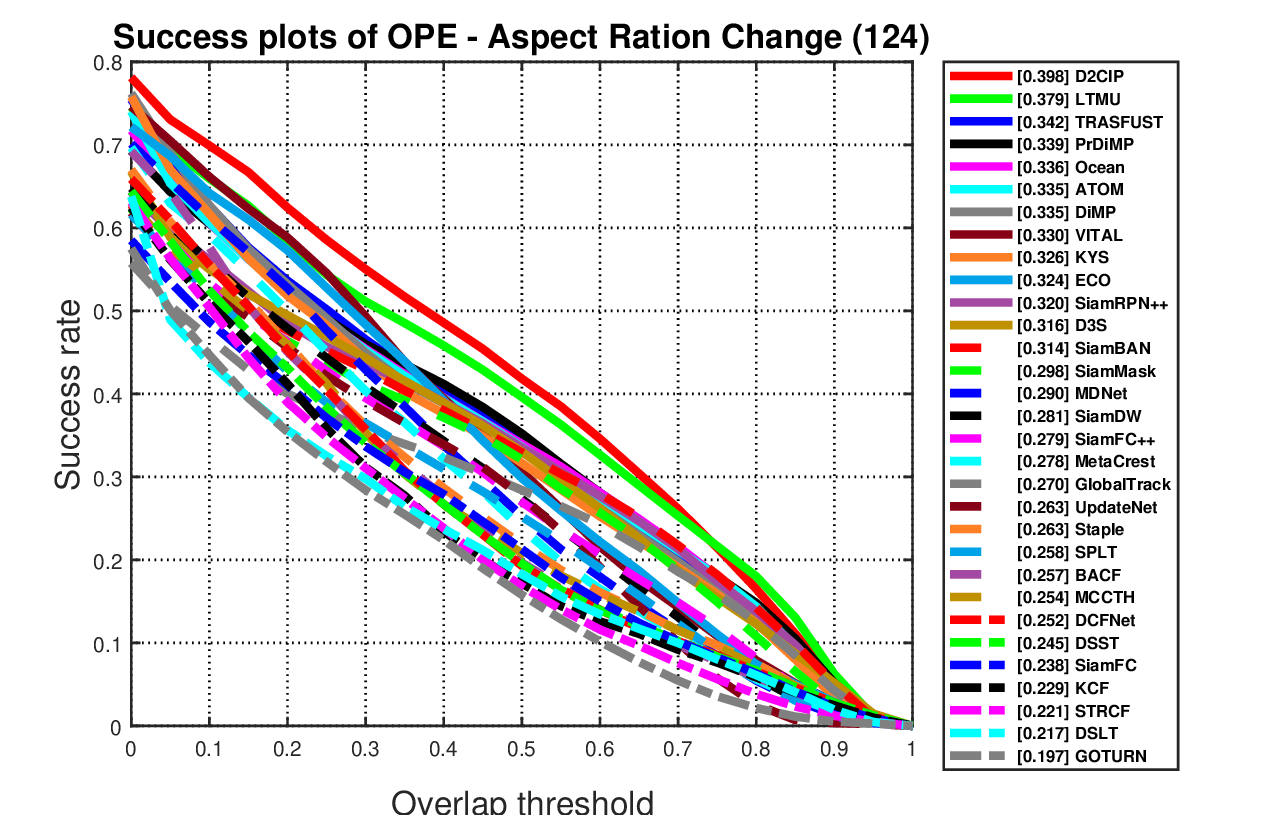}
\includegraphics[trim=30 2 30 2, clip, width=.27\textwidth]{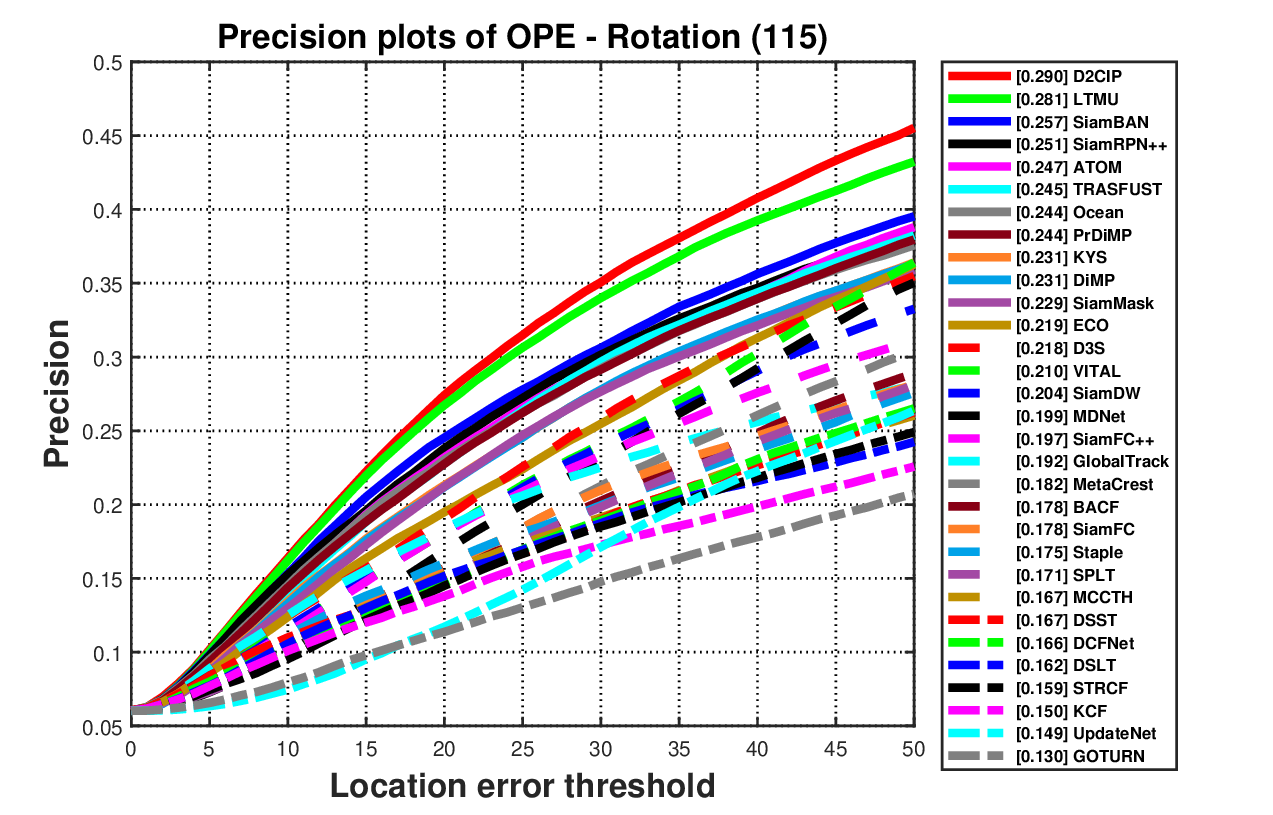}
\includegraphics[trim=30 2 30 2, clip, width=.27\textwidth]{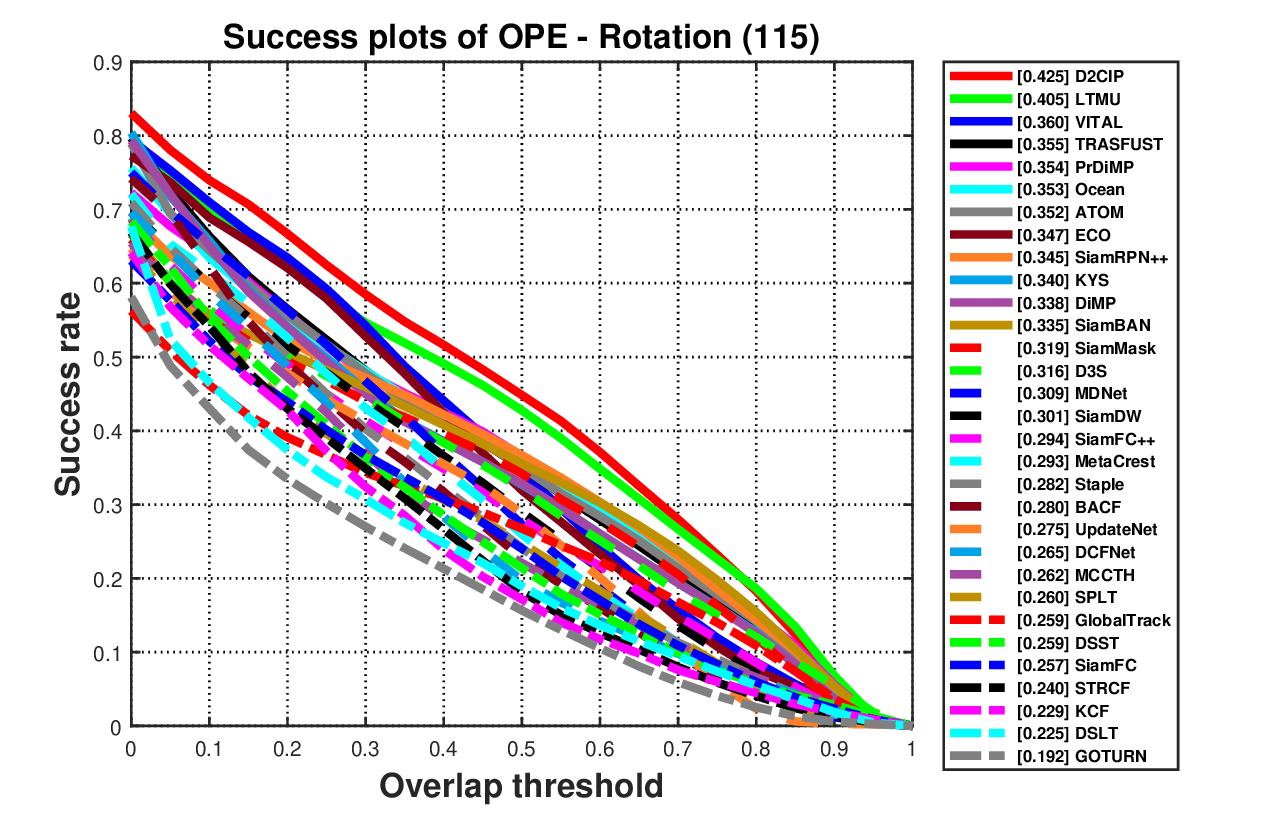}
\centering
\begin{small}
\caption{Quantitative assessment of the performance of our tracker in comparison with state-of-the-art trackers using a one-pass evaluation on the TREK-150 benchmark dataset.}
\label{fig:OPETREK}
\end{small}
\end{minipage}   
\end{figure*}

\subsection{TREK-150 evaluation}
TREK-150 \citep{TREK150} is a recent visual tracking benchmark dataset that includes 150 densely annotated video sequences obtained from a first person perspective. Because no existing visual tracker has used this benchmark in its training process, it is a valuable resource to perform an accurate comparative evaluation of different tracking algorithms. 
Fig. \ref{fig:OPETREK} presents the OPE results of our algorithm on the TREK-150 dataset in comparison with $31$ state-of-the-art trackers including methods based on deep Siamese networks such as SiamFC++ \citep{SiamFC++}, SiamBAN \citep{SiamBAN}, Ocean \citep{Ocean}, SiamMask \citep{SiamMask}, SiamRPN++ \citep{SiamRPN}, SiamDW \citep{SiamDW}, UpdateNet \citep{UpdateNet}, DSLT \citep{DSLT}, SiamFC \citep{SiamFC}, and GOTURN \citep{GOTURN}, as well as correlation trackers (PrDiMP \citep{PrDiMP}, KYS \citep{KYS}, ECO, ATOM \citep{ATOM}, DiMP \citep{DiMP}), DSST \citep{DSST}, KCF \citep{KCF}, Staple \citep{Staple}, BACF, DCFNet \citep{DCFnet}, STRCF \citep{STRCF2018}, MCCTH \citep{MCCTH}, MOSSE \citep{MOSSE}), tracking-by-detection methods (MDNet, VITAL \citep{Vital}), and trackers based on target segmentation representations (D3S \citep{D3S}), meta-learning (MetaCrest \citep{MetaCrest}), fusion strategies (TRASFUST \citep{TRASFUST}), or long-term tracking mechanisms (SPLT \citep{SPLT}, GlobalTrack \citep{GlOBALtRACK}, and LTMU \citep{LTMU}). 
As seen in Fig. \ref{fig:OPETREK}, our tracker outperforms all the other trackers. In particular, it outperforms the state-of-the-art LTMU by $1.27\%$ and $2.4\%$ in terms of overall precision and success, respectively. It is worth noting that our method outperforms SiamBAN, the third best tracker in terms of success by a significant margin ($7.38\%$ relative improvement). As seen in Fig. \ref{fig:OPETREK}, our improvement in comparison to LTMU is higher than $6.5\%$ in attribute "Fast Motion". Furthermore, the long-term meta-update strategy proposed in LTMU could  be integrated with our method, which would likely lead to further performance improvements.

\subsection{OTB100 evaluation}

\begin{figure*}[t]
     \centering 
\begin{minipage}{1\linewidth}
\centering
\includegraphics[trim=10 2 30 2, clip, width=.27\textwidth]{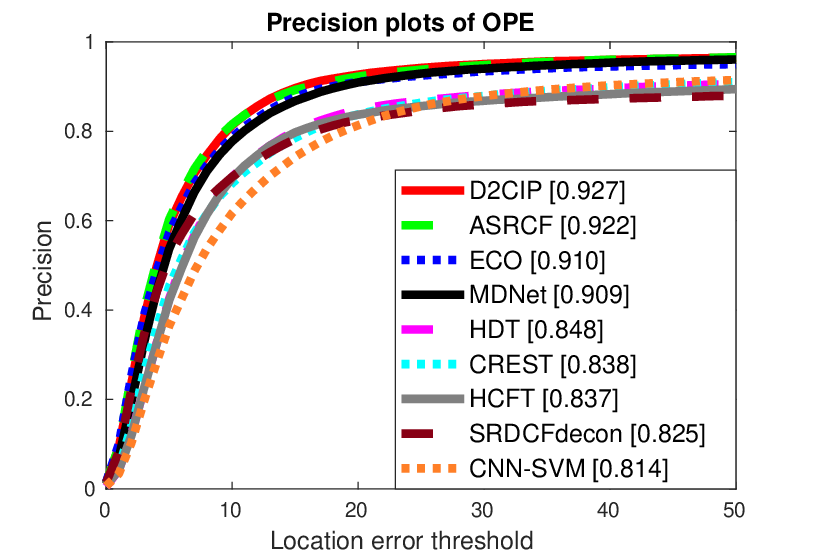}
\includegraphics[trim=10 2 30 2, clip, width=.27\textwidth]{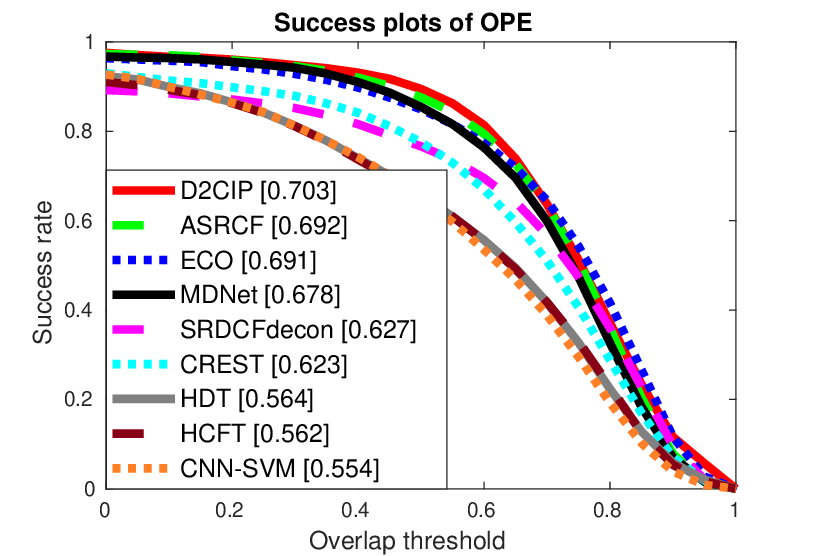}
\includegraphics[trim=10 2 30 2, clip, width=.27\textwidth]{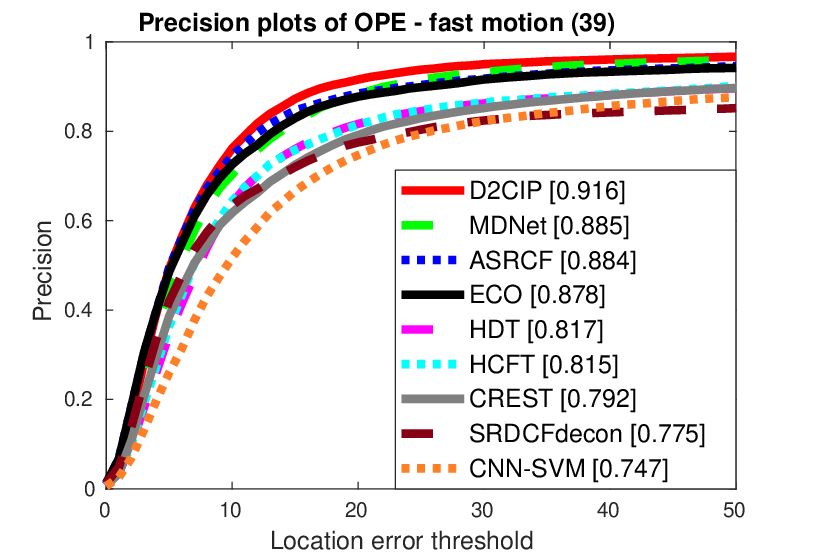}
\end{minipage}  
\begin{minipage}{1\linewidth}
\centering
\includegraphics[trim=10 2 30 2, clip, width=.27\textwidth]{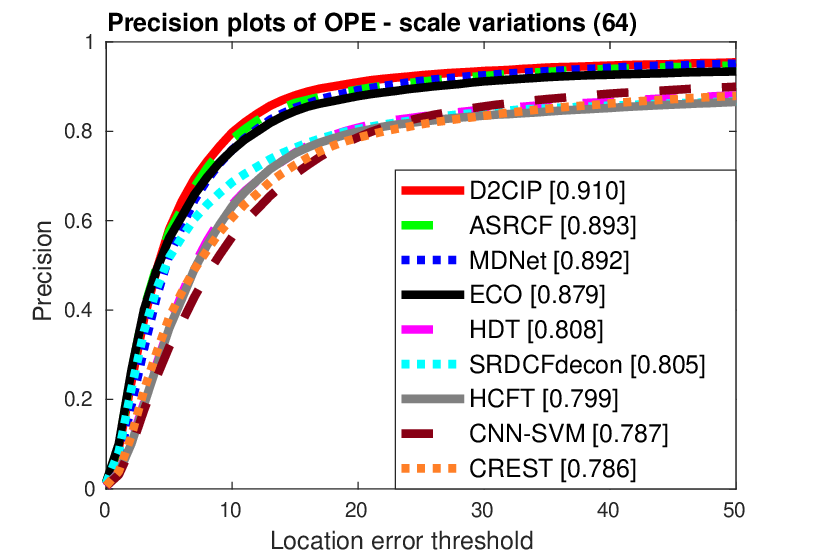}
\includegraphics[trim=10 2 30 2, clip, width=.27\textwidth]{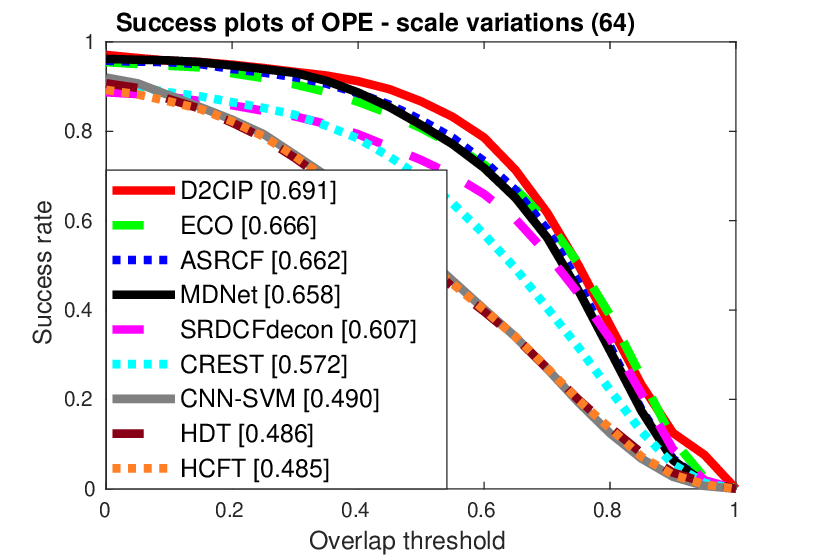}
\includegraphics[trim=10 2 30 2, clip, width=.27\textwidth]{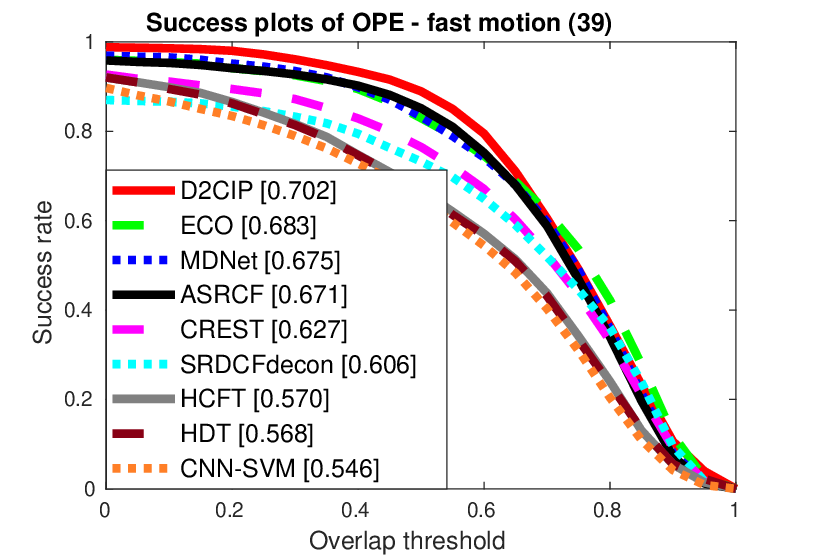}
\end{minipage}  
\begin{minipage}{1\linewidth}
\centering
\includegraphics[trim=10 2 30 2, clip, width=.27\textwidth]{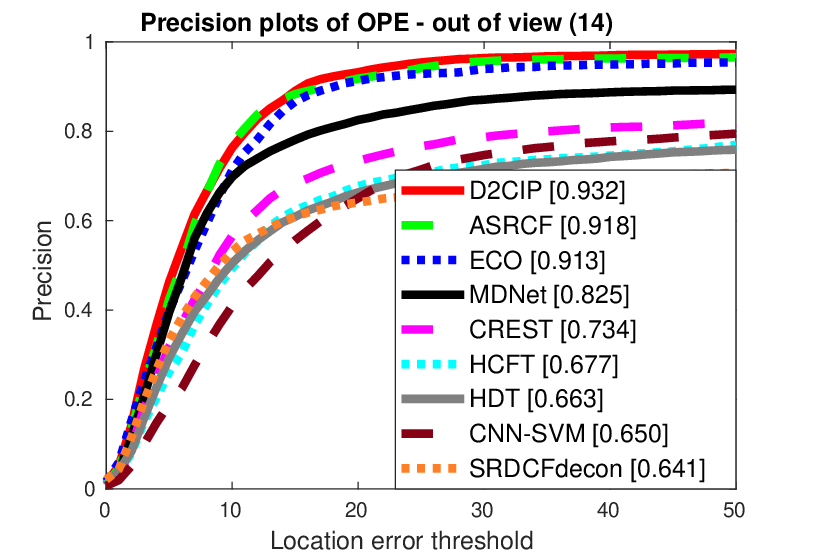}
\includegraphics[trim=10 2 30 2, clip, width=.27\textwidth]{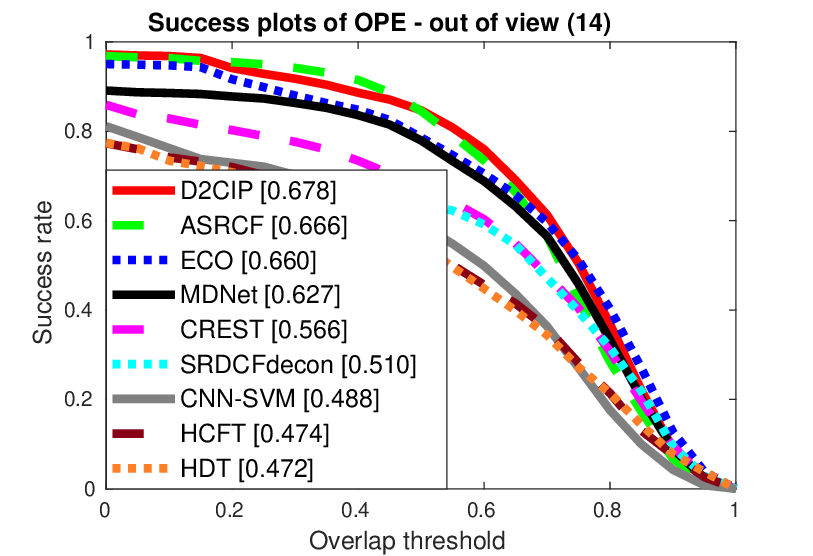}
\includegraphics[trim=10 2 30 2, clip, width=.27\textwidth]{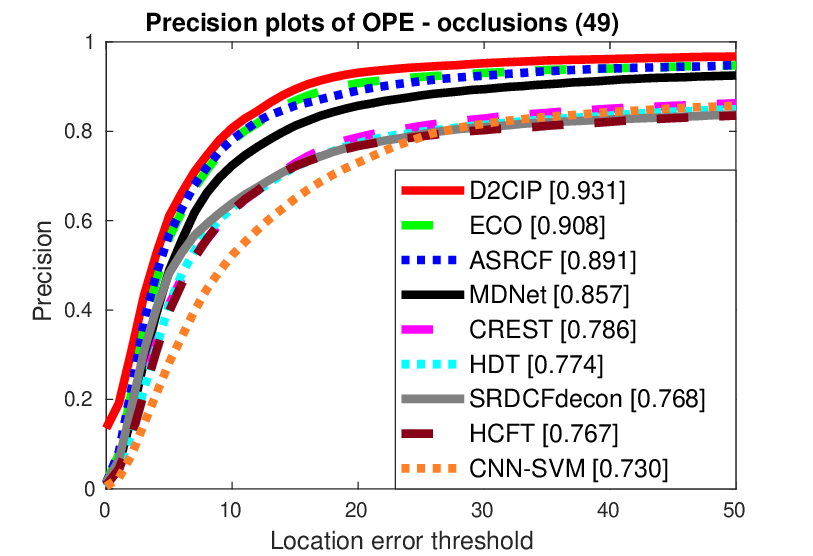}
\end{minipage}  
\begin{minipage}{1\linewidth}
\centering
\includegraphics[trim=10 2 30 2, clip, width=.27\textwidth]{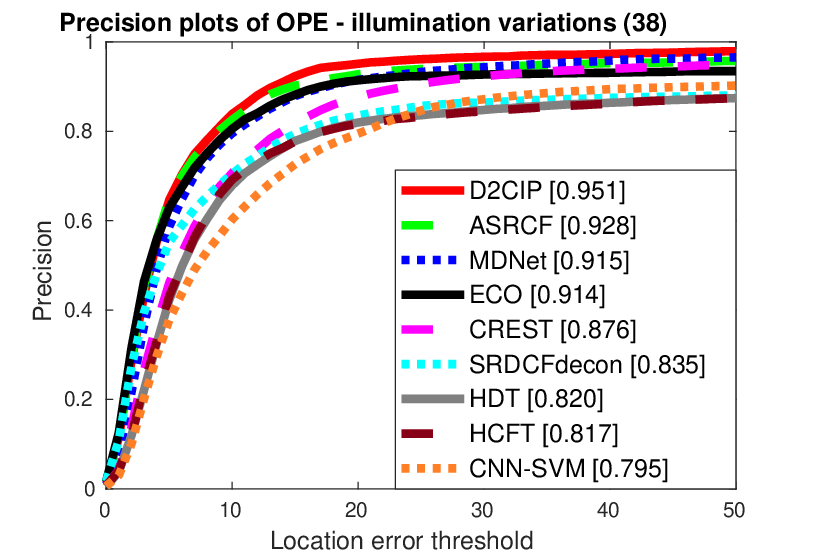}
\includegraphics[trim=10 2 30 2, clip, width=.27\textwidth]{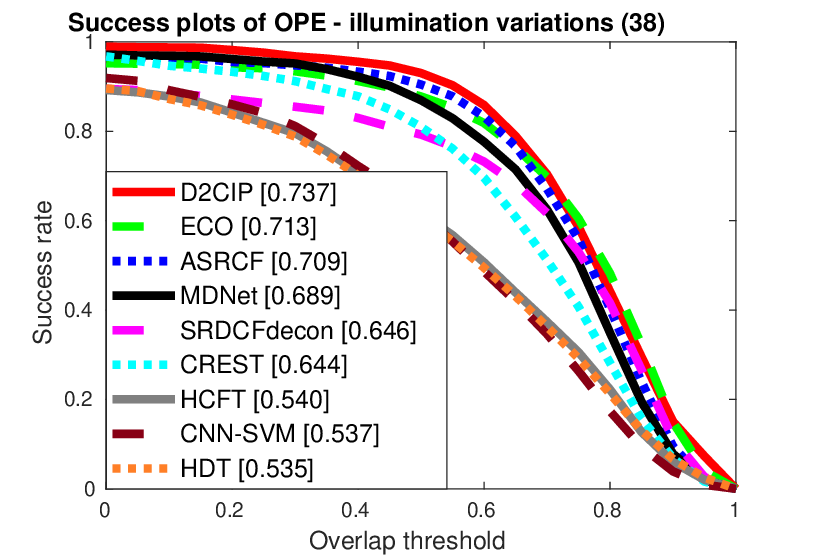}
\includegraphics[trim=10 2 30 2, clip, width=.27\textwidth]{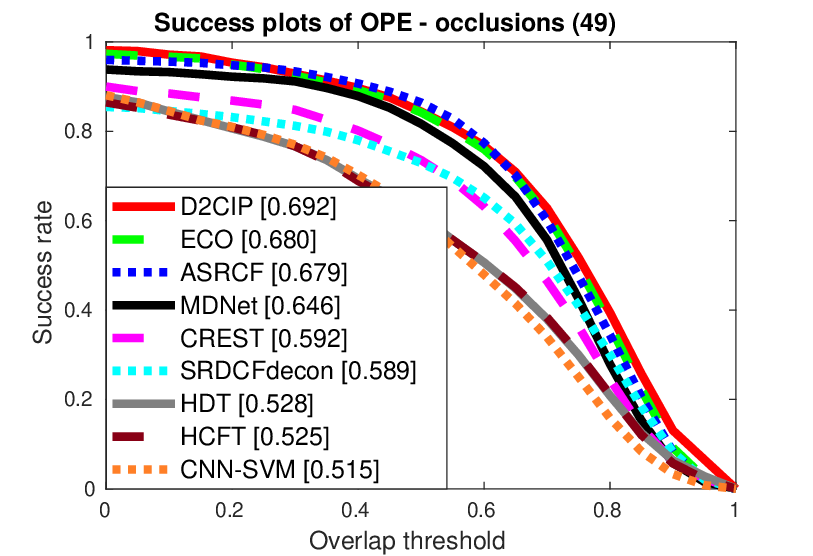}
\begin{small}
\caption{Quantitative performance assessment of our tracker in comparison with eight state-of-the-art trackers using a one-pass evaluation on the OTB100 benchmark dataset.}
\label{fig:OPE}
\end{small}
\end{minipage}  
\end{figure*}

The OTB100 benchmark contains 100 data sequences that are annotated with $11$ attributes. In Fig. \ref{fig:OPE}, we provide a quantitative OPE assessment of our proposed approach in comparison with eight state-of-the-art trackers whose results in the OTB100 dataset are publicly available: ASRCF, ECO, MDNet \citep{MDNet2016CVPR}, HDT \citep{qi2016hedged}, HCFT, FRDCFdecon \citep{DBLP:journals/DanelljanHKF16a}, CREST \citep{song-iccv17-CREST} and CNN-SVM \citep{CNNSVMICML2015}. On the overall evaluation of the precision and success metrics considering the entire dataset, our tracker shows improvements of approximately $0.6\%$ and $1.6\%$ in comparison with the second best tracker ASRCF. Our precision and success improvements in comparison with ASRCF reach $3.6\%$ and $4.6\%$ in fast motion scenarios, $1.9\%$ and $4.4\%$ in scale variation scenarios, $1.5\%$ and $1.8\%$ in out of view conditions, $2.5\%$ and $3.9\%$ when illumination variations are present and $4.4\%$ and $1.9\%$ in scenes including occlusion. In some challenging scenarios in the OTB100 dataset, ASRCF is outperformed by ECO and MDNet. ECO is the second best tracker in the success metric for fast motion, scale variation and illumination variation scenarios as well as both metrics for occlusion scenarios. Our performance improvements with respect to ECO in these scenarios are $2.8\%$, $3.8\%$, $3.4\%$, $2.5\%$, and $1.8\%$, respectively. MDNet only outperforms ASRCF in the precision metric for fast motion scenarios. In that case, our performance improvement with respect to MDNet is $3.5\%$.

\begin{figure*}[t]
     \centering
\begin{minipage}{1\linewidth}
\centering
\includegraphics[width=.8\textwidth]{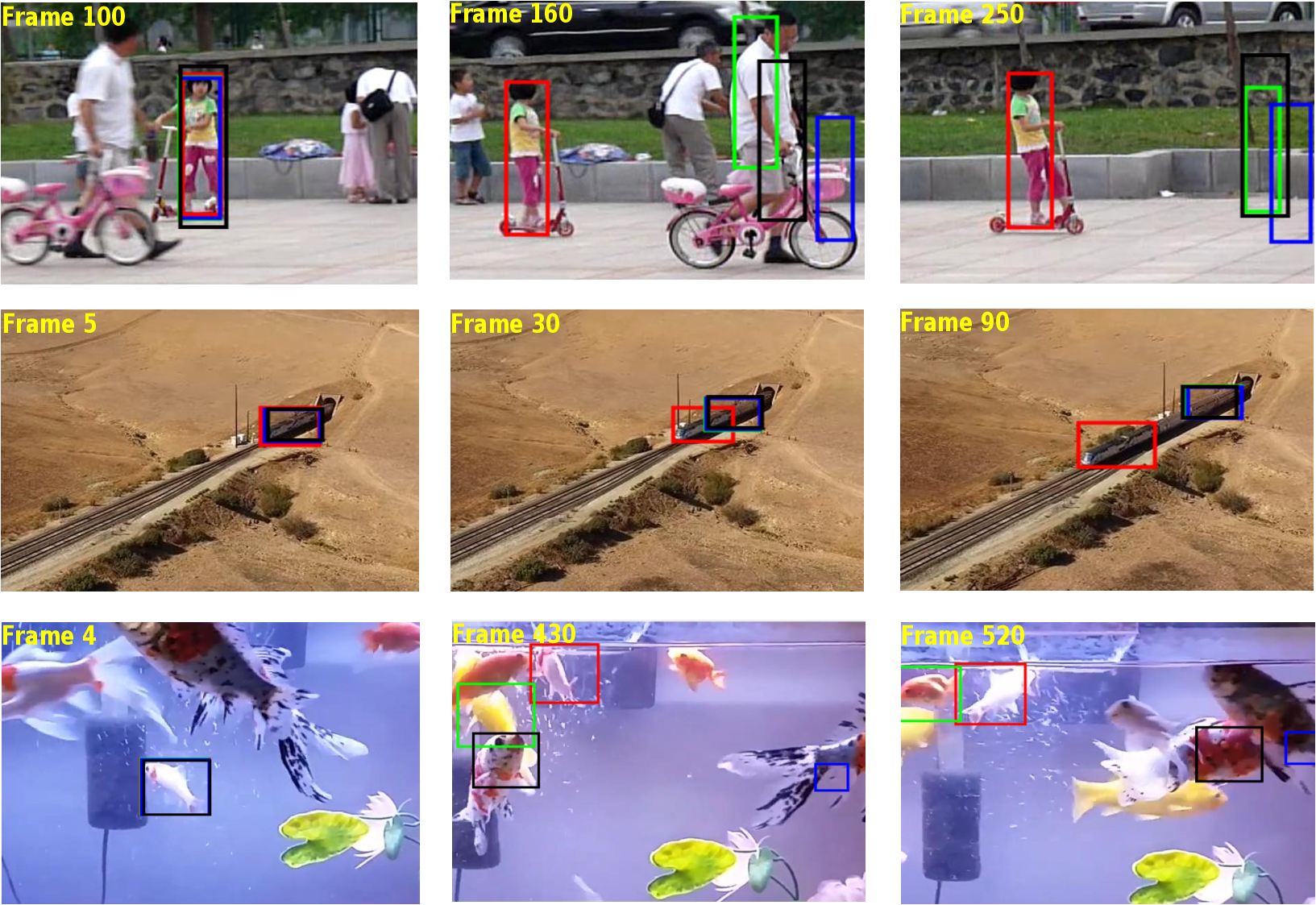}
\end{minipage} 
\begin{tikzpicture}
 \draw [line width=0.8mm, red,left] (0,0) -- (.5,0) node 
 [right,color=black] (text1) {Ours};;
 \draw [line width=0.8mm, green] (text1.east) -- 
 ([xshift=5mm]text1.east) node [right,color=black] (text2) {ASRCF};;
 \draw [line width=0.8mm, blue] (text2.east) -- 
 ([xshift=5mm]text2.east) node [right,color=black] (text3) {ECO};;
 \draw [line width=0.8mm, black] (text3.east) -- 
 ([xshift=5mm]text3.east) node [right,color=black] (text4) {HCFT};;
\end{tikzpicture}
\begin{small}
\caption{Qualitative evaluation of our tracker in comparison with ASRCF, ECO and HCFT on three challenging sequences of the OTB100 (top row, \textit{Girl2} sequence) and LaSOT datasets (middle and bottom rows, \textit{Train-1} and \textit{Goldfish-4} sequences, respectively).}
\label{fig:OPE2}
\end{small}
\end{figure*}
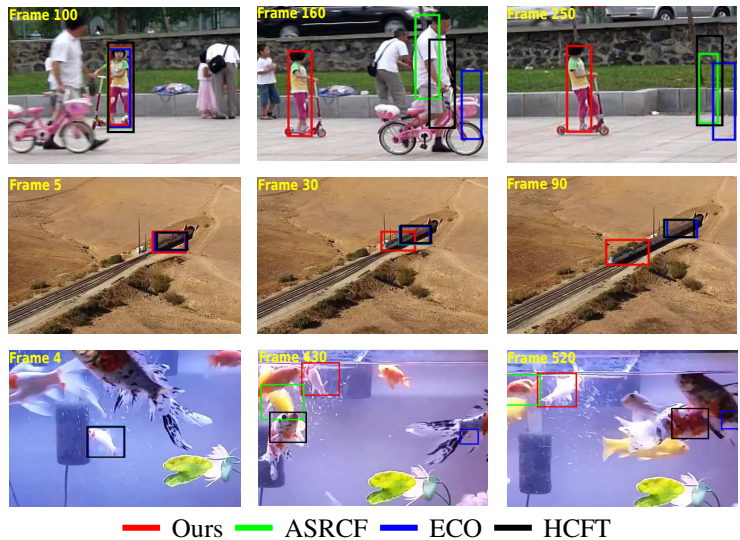

\begin{table*}[t!]
\centering
\caption{Ablative analysis of the components of the proposed visual tracker. PF corresponds to the integration of the baseline tracker with the particle filter model. IPF incorporates the iterative particle refinement procedure. IPFK includes the clustering of the likelihood distributions. D2CIP corresponds to our complete algorithm, which further incorporates maintaining one target model for each predicted mode. The numbers within parentheses indicate the relative performance gains over the baseline tracker.}
\label{Improvement}
\setlength\tabcolsep{4.5pt} 
\begin{tabular}{cccccccc}
\hline
Benchmark                    & PF & IPF & IPFK & D2CIP      & Precision  & Success \\
\hline
\multirow{4}{*}{LaSOT}     &\checkmark  &\xmark &\xmark  &\xmark      &$30.9\%$ ~($+2.6\%$)  &$33.1\%$ ~($+2.3\%$)  \\ 
                           &\checkmark  &\checkmark &\xmark  &\xmark  &$32.3\%$ ~($+7.3\%$)  &$34.5\%$ ~($+6.4\%$)  \\
                       &\checkmark  &\checkmark &\checkmark  &\xmark  &$33.2\%$ ($+10.4\%$)  &$34.9\%$ ~($+7.9\%$)  \\
                &\checkmark  &\checkmark &\checkmark  &\checkmark   &$33.5\%$ ($+11.3\%$)  &$35.2\%$ ($+8.7\%$) \\
                     \hline 
\multirow{1}{*}{TREK-150}    
                &\checkmark  &\checkmark &\checkmark  &\checkmark   &$32.0\%$ ($+18.96\%$)  &$41.9\%$ ($+15.11\%$) \\
                     \hline                      
\multirow{4}{*}{OTB100}   &\checkmark  &\xmark &\xmark  &\xmark  &$91.5\%$ ~($+0.5\%$)  &$69.6\%$ ~~($+0.7\%$) \\ 
                         &\checkmark  &\checkmark &\xmark  &\xmark  &$92.2\%$ ~($+1.3\%$)  &$69.9\%$ ~($+1.2\%$)  \\
                    &\checkmark  &\checkmark &\checkmark  &\xmark   &$92.6\%$ ~($+1.8\%$)  &$70.2\%$ ~($+1.6\%$)  \\
                &\checkmark  &\checkmark &\checkmark  &\checkmark   &$92.7\%$ ~($+1.9\%$)  &$70.3\%$ ~($+1.7\%$) \\
\hline
\end{tabular}
\end{table*}

Fig. \ref{fig:OPE2} presents a qualitative assessment of our tracker in comparison with ASRCF, ECO, and HCFT. In the first row, the other trackers fail because of a relatively long occlusion period, which causes not only tracking loss but also  incorrect model updates. Our tracker, on the other hand, finds multiple potential clusters during partial occlusion and maintains one model per cluster. When the target becomes visible again, this enables our tracker to sample from distributions closer to the target and assess the corresponding updated models. In the second row, the other trackers fail because of a period of fast motion by the target. Our particle filter enables our tracker to sample particles over a wider search area and the iterative particle refinement allows it to reach the best location for the target. In the third row, the presence of several similar objects causes failures in the other trackers, whereas our novel method for evaluating the particle likelihoods locates the correct target accurately.

\subsection{Ablative analysis and computation time}
Table \ref{Improvement} shows the performance improvement introduced by the different components of our proposed method and compares them with ECO \citep{ECO}, our baseline tracker. As the table indicates, the incorporation of a simple constant velocity particle filter with the baseline tracker leads to relative improvements of $2.6\%$ and $2.3\%$ in terms of precision and success on the LaSOT dataset. The iterative particle refinement procedure further improve these results by $4.7\%$ and $4.1\%$. Clustering the particles to identify the modes of the likelihood provide additional gains of $3.1\%$ and $1.5\%$. Finally, maintaining one target model for each mode of the distribution leads to relative improvements for $0.9\%$ and $0.8\%$. That is, our proposed method improves the overall performance of ECO by $11.3\%$ in terms of precision and $8.7\%$ on the success metric for LaSOT. 

Since the OTB100 dataset contains less complex sequences, the performance of the baseline tracker is already relatively high for both metrics. For example, one of the main benefits of the iterative particle refinement procedure is that is generates more precise target estimates, reducing model drift. While this is easily observed in the longer videos comprising the LaSOT dataset, there are significantly fewer opportunities to correct model drift in the shorter videos contained in the OTB100 dataset. In addition, since the OTB100 dataset includes only $16$ target categories, the convolutional features generated by the backbone CNN are much more discriminative than for the $85$ categories present in LaSOT. As a consequence, the robustness to background clutter introduced by the multi-modal likelihood model are dramatically less pronounced in OTB100. Nonetheless, our tracker still provides up to $1.9\%$ and $1.7\%$ relative improvements in precision and success.

Our unoptimized MATLAB implementation of D2CIP runs at approximately 2 FPS on an RTX 2080Ti GPU. The main bottleneck in our current implementation is the computation of the convolutional features used in our likelihood model, which accounts for more than $80\%$ of the processing time. However, since there are no inter-dependencies among particles, it is possible to compute their likelihoods in parallel. We have observed that the hardware resources are severely underutilized, which indicates room for reduction in the overall computation time. 

\section{Conclusion}
This work proposes an iterative particle filter that works along with a deep convolutional neural network and a correlation filter to accurately track objects of interest in video sequences. In the proposed algorithm, after generating a set of initial particles around the predicted target sizes and positions and extracting their hierarchical convolutional features, a correlation map is  calculated for each particle. These maps are used to refine the positions of the particles and generate the corresponding likelihoods. After discarding low likelihood particles, the displacement between each particle and the peak of its correlation response map is calculated. If the distance between the peak of the correlation response map and the particle location is larger than a threshold $\epsilon$, the image patch centered at the peak of the response map is used to generate a new response map. This iterative procedure leads most particles to converge to only a few final positions. By iteratively updating the particle likelihoods, our method also addresses the problem of calculating the posterior distribution over the correct support points in particle-correlation trackers. Finally, a novel method is used to assess multi-modal likelihoods based on clustering the particles. The LaSOT, TREK-150, and OTB100 datasets are used for evaluating the proposed tracker's performance. The results show that our tracker substantially outperforms several state-of-the-art methods.

\section*{Acknowledgment}

This material is based upon work supported by the National Science Foundation under Grant No. 2224591.

\bibliography{refs}

\end{document}